\documentclass[journal]{IEEEtran}

\usepackage{bm}
\usepackage{amsmath}
\usepackage{epsf}
\usepackage{graphics}
\usepackage{ amssymb }
\usepackage[dvips]{graphicx}
\usepackage{mathtools}% Loads amsmath
\usepackage{epsfig}
\usepackage{cite}
\usepackage[linesnumbered,ruled,vlined]{algorithm2e}
\usepackage{colortbl}
\usepackage{color}
\usepackage{enumitem}
\usepackage{makecell}

\usepackage{bm}
\usepackage{amsmath}
\usepackage{epsf}
\usepackage{graphics}
\usepackage{ amssymb }
\usepackage[dvips]{graphicx}
\usepackage{epsfig}
\usepackage{cite}
\usepackage[linesnumbered,ruled,vlined]{algorithm2e}
\usepackage{graphicx}
\usepackage{epsfig}
\usepackage{latexsym}
\usepackage{amsfonts}
\usepackage{here}
\usepackage{rawfonts}

\usepackage[utf8]{inputenc}
\usepackage[english]{babel}
\usepackage{amsmath}
\usepackage{amsfonts}
\usepackage{amssymb}
\usepackage{color} %May be necessary if you want to color links
\usepackage{bm}
\usepackage{listings}
\usepackage{caption}
    \usepackage{multicol}

\usepackage{tabularx,booktabs}
\newcolumntype{Y}{>{\centering\arraybackslash}X}
\makeatletter
\newcommand\notsotiny{\@setfontsize\notsotiny{6.31415}{7.1828}}
\makeatother
\usepackage{amssymb}
\usepackage{amsthm}
\usepackage{graphicx}
\usepackage{epstopdf}
\usepackage{listings}
\usepackage{float}
\usepackage{amsmath}
\usepackage{amssymb}
\usepackage{amsfonts}
\usepackage{epstopdf}

\usepackage{multirow}
\usepackage{amscd}
\usepackage{mathrsfs}
\usepackage{graphicx}
\usepackage{color}
\usepackage{url}
\usepackage{bm}
\usepackage{ulem} %for editing purposes only
\usepackage{setspace}
\usepackage{footnote}
\DeclareMathOperator*{\argmin}{arg\,min}

\newtheorem{remark}{Remark}

\usepackage[]{algorithm2e}

\addto\captionsenglish{}
\usepackage{bbm}

\usepackage{multicol}
\usepackage{mathtools}

\usepackage{lipsum,graphicx}
\usepackage{subcaption}

\usepackage{diagbox}
\usepackage{hyperref}
\usepackage[protrusion=true,expansion=true]{microtype}
\pdfoutput=1
\usepackage[font=footnotesize]{caption}
\allowdisplaybreaks[4]

\usepackage{graphicx}
\usepackage{grffile}
\usepackage{tabularx}
\newcounter{term}[section]

\renewcommand\theterm{\alph{term}}
\makeatletter
\newcommand{\vast}{\bBigg@{4}}

\newcommand{\Vast}{\bBigg@{5}}
\makeatother

\usepackage{soul}

\usepackage[most]{tcolorbox}
\usepackage{color}
\definecolor{Gray}{gray}{0.9}

\definecolor{LightBlue}{rgb}{0.8,0.89,1}
\definecolor{DarkBlue}{rgb}{0.47,0.784,1}
\definecolor{MagLight}{rgb}{1, 0.89, 0.8}
\newtagform{blue}{\color{blue}(}{)}
\usepackage{hhline}
\newcolumntype{?}{!{\vrule width 1pt}}
\newlength{\rowhgtspace} 
\newcommand{\rowspacer}[1]{\begin{tabular}{@{}c@{}} #1 \end{tabular}}

\tcbset{colback=yellow!10!white, colframe=black!50!black, 
        highlight math style= {enhanced, %<-- needed for the ’remember’ options
            colframe=black,colback=red!10!white,boxsep=0pt}
        }
        \allowdisplaybreaks

\begin{document} 

\title{Multi-Modal Federated Learning for Cancer Staging over Non-IID Datasets with Unbalanced Modalities\vspace{-1mm}}
\author{Kasra Borazjani,~\IEEEmembership{Student Member,~IEEE}, Naji Khosravan,~\IEEEmembership{Member,~IEEE}, \\Leslie Ying,~\IEEEmembership{Senior Member,~IEEE}, and Seyyedali Hosseinalipour,~\IEEEmembership{Member,~IEEE}
\vspace{-3mm}
\vspace{-3mm}
\thanks{K. Borazajani and S. Hosseinalipour are with the Department of Electrical Engineering, University at Buffalo--SUNY, NY, USA (Email:\{kasrabor,alipour\}@buffalo.edu). N. Khosravan is with Zillow Group~\textregistered, WA, USA (Email:najik@zillowgroup.com). L. Ying
is with the Department of Biomedical Engineering and Electrical Engineering, University at Buffalo--SUNY, NY, USA (Email:leiying@buffalo.edu).

}

}
\maketitle
% \vspace{-25mm}
\setulcolor{red}
\setul{red}{2pt}
\setstcolor{red}
\setlength{\abovedisplayskip}{3.5pt}
\setlength{\belowdisplayskip}{3.5pt}

\begin{abstract}

The use of machine learning (ML) for cancer staging through medical image analysis has gained substantial interest across medical disciplines. When accompanied by the innovative federated learning (FL) framework, ML techniques can further overcome privacy concerns related to patient data exposure.
Given the frequent presence of diverse data modalities within patient records, leveraging FL in a multi-modal learning framework holds considerable promise for cancer staging.
 However, existing works on multi-modal FL often presume that all data-collecting institutions have access to all data modalities. This oversimplified approach neglects institutions that have   access to only a portion of data modalities within the system.
In this work, we introduce a novel FL architecture designed to accommodate not only the heterogeneity of data samples, but also the inherent heterogeneity/non-uniformity of data modalities across institutions. We shed light on the challenges associated with varying convergence speeds observed across different data modalities within our FL system. Subsequently, we propose a solution to tackle these challenges by devising a distributed gradient blending and proximity-aware client weighting strategy tailored for multi-modal FL.
To show the superiority of our method, we conduct experiments using \textit{The Cancer Genome Atlas program (TCGA)} datalake considering different cancer types and three modalities of data: mRNA sequences, histopathological image data, and clinical information. Our results further unveil the impact and severity of \textit{class-based} vs \textit{type-based} heterogeneity across institutions on the model performance, which widens the perspective to the notion of \textit{data heterogeneity} in multi-modal FL literature.
\end{abstract}
\vspace{-1mm}

\vspace{-2mm}

\section{Introduction}
\label{introduction}

\noindent
Leveraging medical images to derive machine learning (ML) models for aiding in cancer staging and developing effective treatment strategies has shown significant success~\cite{centralSingleModal1}. 
Traditionally, despite the existence of multiple modalities of data  (e.g., computed tomography (CT) scans \cite{sarkar2023,detect-lc-22} and histopathological images \cite{Xiehist19}), ML models (e.g., MLPs, CNNs, and LSTMs) are trained with one of data modalities~\cite{litjens2016deep, hartenstein2020prostate, hadjiyski2020kidney}.
However, as research revealed the benefits of multi-modal ML~\cite{huang2021makes}, modern approaches have utilized multiple data types/modalities such as mRNA sequences and clinical information together to obtain ML models for medical applications~\cite{SHAO2020101795}.

The majority of existing studies on
 multi-modal ML within the medical domain \cite{venugopalan2021multimodal,sun2018multimodal,xu2020accurately} presume a \textit{centralized ML model training architecture}: ML models are trained at a module (e.g., a server) with access to a repository of patient data.
However, since patient
data is often distributed across various data collection units
(e.g., hospitals, clinics), the centralized approach
necessitates the transfer of patient data across the network
to a centralized location, raising data privacy
concerns.
Researchers have thus explored alternative ML approaches, a notable example of which is federated learning
(FL)~\cite{pmlr-v54-mcmahan17a, rieke2020future}.

In FL, each data collecting institution, referred to as a \textit{client}, independently trains a local ML model using its data. These local ML models are periodically aggregated by a server to construct a \textit{global model}. The global model is disseminated to clients to commence their next local ML model training cycle.
Consequently, FL replaces the transmission of raw data across the network with the transmission of ML models, thus preserving data privacy.
As a result, the adoption of FL within the medical domain is gaining substantial traction \cite{fed-med-lit-rev, fedpatientsim, qayyum2022collaborative, park2021federated, feng2022specificity}.
Given the diversity of data  types/modalities across FL clients in medical domain, such as hospitals and clinics, the prospect of employing FL within a multi-modal learning context is particularly enticing. Although FL and multi-modal ML have been jointly studied recently \cite{qayyum2022collaborative, CREMONESI2023104338}, a common constraint in these studies is the assumption that each client has access to all data modalities.
In real-world scenarios, this assumption often does not hold: for example, a hospital may collect mRNA sequences and histopathological images, while a clinic may collect only CT scans and clinical information. 

In this work, we take an initial step towards implementing multi-modal FL in a scenario where  clients not only have non-iid (non-independent and identically distributed) data, but also possess an uneven number of data modalities. We present a system model that eliminates the need for clients to possess identical sets of data modalities by employing a versatile distributed encoder-classifier architecture.
Within this framework, the encoders are tasked with the extraction of features from each distinct data modality at each client. These encoders and classifiers/decoders will undergo separate aggregations in order to address the heterogeneity of modalities across clients. This approach ensures that all trainable components relevant to a specific data modality are collectively trained using the data of that particular modality across all clients. The final global model resulting from this training can then be applied over the combination of all data modalities.

We show that while our proposed system model holds considerable promise, it confronts a series of challenges. A notable concern is the disparity in the training and convergence speeds of various data modalities among different clients. As we will later show (Sec.~\ref{subsec:key-challenges}), based on the combination of the modalities available in a specific client's data, each modality can converge at a different speed. This variability can lead to certain encoders rapidly exhibiting a biased behavior towards local data of  clients, which can impede the global model's convergence.\footnote{This phenomenon, in turn, may result in overfitting toward specific data labels, particularly in the context of generic FL classification tasks,  due to the imbalanced effect of each modality during local ML training.}
To tackle this challenge, we introduce a novel approach termed \textit{distributed gradient blending} (DGB) for FL. Although conceptually distinct, this approach draws inspiration from a seminal work on centralized multi-model ML~\cite{gradient-blending}. 
In a nutshell, DGB capitalizes on information pertaining to the performance of participating clients in FL and the characteristics of their held data modalities to determine effective back-propagation weights for local training of each encoder-classifier pair at clients. In particular, in DGB, the back-propagation weights (i.e., the learning rate that tunes the sensitivity to the obtained gradients during the training process) is tuned based on the growth in the performance of the modalities in terms of their generalization and overfitting behavior. Nevertheless, DGB by itself cannot take into account for the heterogeneity (i.e., non-iid-ness) of data distributions\footnote{In this paper, heterogeneity of data distribution is captured via both \textit{class-based heterogeneity} (e.g., each client can have different ratio of data points belonging to different stages of cancer, such as stage I and II, in its dataset) and \textit{type-based heterogeneity} (e.g., clients possessing data of various cancer types, such as breast, lung, and liver cancers).} across clients, which is a unique challenge in distributed networks of interest to FL. Subsequently, to enhance the robustness of our proposed method against non-iid data distribution across the clients, we propose \textit{proximity-aware client weighting} (PCW). In a nutshell, PCW further tunes the back-propagation weights in DGB via explicitly taking into account for the quality of data at  each client, measured via the similarity of the cumulative gradient of the local ML model of the client and that of the global model.

The major contributions of this work are summarized below:
\begin{itemize}[leftmargin=4.5mm]
    \item We present a novel system model that accommodates an uneven number of data modalities across clients in FL. This introduces an additional layer of data heterogeneity, superimposed upon the inherent non-iid nature of per-modality data distribution across the clients in FL. The introduction of this system model allows real-world applications of multi-modal FL, particularly within the domain of healthcare.
    \item We identify challenges associated with achieving high-performance ML models within our system. To tackle a crucial challenge related to disparate convergence rates across data modalities of various clients, we introduce DGB. This approach is specifically designed for FL systems with clients having uneven numbers of data modalities. DGB harnesses system-wide information, namely, the training and validation losses, and data characteristics, to dynamically adjust the learning-rate of each modality  at each FL client.
    \item To enhance the convergence speed and thus improving the communication efficiency of our approach, we introduce PCW. PCW  
regulates the calculation of DGB parameters with respect to the varying qualities of data present across the FL clients (i.e., the fidelity of estimate). 
    \item We conduct numerical evaluations on a dataset of three cohorts gathered by The Cancer Genome Atlas program (TCGA - \href{https://www.cancer.gov/tcga}{https://www.cancer.gov/tcga}) for three different cancer types (breast, lung, and liver) and demonstrate that our method can lead to significant performance gains as compared to various baseline methods in multi-modal FL.
    In our experiments, we further study the impact of \textit{class-based} vs \textit{type-based} heterogeneity across institutions on the model training performance. This study opens a new perspective to the notion of \textit{data heterogeneity} not only in FL literature focused on downstream task of cancer staging but also in the broader multi-modal FL literature.
\end{itemize}

\vspace{-4mm}

\section{Related Work}
\label{literature}
\vspace{-0.5mm}

\noindent In this section, we review the existing literature on cancer staging using ML methods and the expansions and contributions of FL to this active research area. Henceforth, we use the terminology of \textit{ML models} to refer to deep neural networks.

\textbf{Single-Modal Cancer Staging.}
In most of the early works \cite{ litjens2016deep, hartenstein2020prostate, hadjiyski2020kidney, detect-lc-22, sarkar2023, Xiehist19}, only one modality of data is used in order to train the ML models. Thus, the ML models developed in these works take an input datapoint of a specific modality (e.g., PET/CT image or histopathological image) and pass it through -- either raw or preprocessed -- to predict its class (e.g., the cancer stage). As a result, most of the recent efforts in this area~ \cite{hartenstein2020prostate,hadjiyski2020kidney}, have been focused on optimizing the ML model architecture and developing effective data preprocessing methodologies. However, the aformentioned studies overlook the proliferation of diverse data modalities that hold the potential to significantly enhance the accuracy of ML models for cancer prediction and staging. Moreover, these studies often make the assumption of unrestricted access to a comprehensive reservoir of patient data, the acquisition of which is becoming increasingly challenging due to the concerns regarding data privacy. In this work, we aim to address both of these issues by proposing multi-modal FL and resolving the challenges it comes with.

\textbf{Multi-Modal Cancer Staging.}
Researchers have investigated the advantages of leveraging diverse data modalities for cancer prognosis \cite{boehm2022multimodal, olatunji2023multimodal, SHAO2020101795, mm-dl-cl-2022, aacr-mRNA-ml}. For instance, in \cite{SHAO2020101795}, the authors demonstrated the significant impact of incorporating histopathological images, gene expression data, and clinical data on the performance of ML models used for cancer staging. Also, the utilization of CT scan data, clinical data, and selected deep-learning-based radiomics data in cancer recurrence prediction has yielded similar ML prediction enhancements in~\cite{mm-dl-cl-2022}. Nevertheless, these studies mostly operate within a centralized ML training architecture, where the ML training module, often hosted on a central server, is assumed to have access to a large-scale repository of patient data. This assumption limits the applicability of these findings in real-world scenarios where patient data is distributed across multiple medical institutions, pooling of which in a centralized location raises major concerns regarding leakage of patients' information. 
Our contribution to this literature lies in the proposition of a distributed ML architecture inspired by FL that eliminates the need for patients data transfer, while explicitly accommodating the diversity of data modalities across distinct data collection institutions.

\textbf{Federated Learning.}
Motivated by the privacy concerns associated with the transfer of data across the network, there has been a notable interest in the application of distributed ML techniques, with particular attention directed towards FL, in various fields of studies \cite{truex2019hybrid, arivazhagan2019federated}. 
Nevertheless, majority of works on FL \cite{yi2020net, sheller2019multi} presume the existence of single modality of data across the clients, and thus cannot be easily  deployed in systems with multiple modalities of data.

\textbf{Multi-modal FL for Cancer Classification.}
Multi-modal FL has recently attracted notable attention within the research community of various fields, such as healthcare and machine vision \cite{qayyum2022collaborative, NANDI2022340, mm-fl-gradblend, XIONG2022110, CREMONESI2023104338, agbley2021multimodal},
among which, considering medical applications, several endeavors stand out \cite{qayyum2022collaborative, agbley2021multimodal}.
In \cite{qayyum2022collaborative}, an FL system effectively incorporates both X-ray and Ultrasound modalities. In \cite{agbley2021multimodal}, multi-modal FL is proposed with skin lesion images and their corresponding clinical data for melanoma detection. However, a common thread runs through all of the previous works, implying both to the general context of multi-modal FL and the cases where it has been applied to medical topics: they all operate under the assumption that every client has access to (and collects) \textit{all} of data modalities.
In practice this assumption often encounters limitations in the medical domain as data collection institutions often gather only a subset of data modalities; for instance, one hospital may gather different gene expression data and clinical data, while another may collect CT scans. Our work is among the first in the literature to introduce an effective FL paradigm across clients with an uneven distribution/number of data modalities.

\vspace{-2mm}
\section{Network and Learning Model}
\label{sec:method}
\noindent In this section, we first present our  network model of interest (Sec. \ref{subsec:network-model}). Subsequently, we offer a concise discussion on conventional uni-modal FL (Sec. \ref{subsec:unimodal-fl}). Our contribution unfolds as we extend the conventional FL paradigm by introducing a novel FL system designed to accommodate a network of clients that possess an uneven distribution/number of data modalities (Sec. \ref{subsec:unbalanced-mmfl}). We conclude this section by providing an overview of the pivotal challenges that necessitate addressing to achieve high-performance ML models within the context of our proposed FL system (Sec. \ref{subsec:key-challenges}). The notations used in the modeling are summarized in Table~\ref{tab:notation-table}.

\vspace{-1.5mm}
\subsection{Network Model} \label{subsec:network-model}
We consider a network of $N$ institutions/clients gathered via the set $\mathcal{N}=\{1, \cdots, N\}$, $|\mathcal{N}|=N$. Each institution $n \in \mathcal{N}$ is assumed to possess a dataset $\mathcal{D}_n = \mathcal{D}_n^{\mathsf{Tr}} \cup \mathcal{D}_n^\mathsf{Va}$ with size $D_n \triangleq |\mathcal{D}_n|$ where $\mathcal{D}_n^{\mathsf{Tr}}$ is the dataset used for \underline{tr}aining and $\mathcal{D}_n^{\mathsf{Va}}$ is the dataset used for \underline{va}lidation. We also denote the union of datasets collected across the institutions, called global dataset, as $\mathcal{D}\triangleq \cup_{n \in\mathcal{N}}{\mathcal{D}_n}$ with $D \triangleq |\mathcal{D}|$ datapoints.\footnote{Without loss of generality, as commonly presumed in the literature of FL \cite{XIONG2022110, CREMONESI2023104338}, we assume that datasets across the institutions are non-overlapping.}

\vspace{-1.5mm}
\subsection{Conventional Uni-modal Federated Learning} \label{subsec:unimodal-fl}
In the conventional uni-modal FL framework, the datasets held by institutions, and consequently the global dataset, consist solely of a single data modality (e.g., image data)~\cite{pmlr-v54-mcmahan17a, truex2019hybrid, arivazhagan2019federated}.  FL conducts ML model training through a series of local (model) training rounds indexed by $k \in \{0,\cdots, K-1\}$, each leading to a global (model) aggregation step, indexed by $t\in\{0,\cdots, T-1\}$.
Within this context, each institution $n$  at global aggregation step $t$ and local training round $k$ trains a local ML model, represented by a parameter vector $\bm{\omega}_n^{(t),k}$.
The initial local training round commences with the server initializing a global model $\bm{\omega}^{(t)}$ at $t=0$ and disseminating it among the institutions. Subsequently, each institution $n$ synchronizes its local model as $\bm{\omega}_n^{(t),k} \leftarrow \bm{\omega}^{(t)}$ for $t=0$ and $k=0$ and utilizes it as the initial model for local training, i.e., $\bm\omega^{(0),0}_n~\leftarrow~\bm\omega^{(0)}$. Following this initialization, each institution undergoes $K$ local stochastic gradient descent (SGD) iterations, utilizing its own dataset to obtain $\bm{\omega}^{(t),K}_n$. The interim ML updates, for $0\leq k \leq K-1$, are computed as $\bm{\omega}^{(t),k+1}_n = \bm{\omega}^{(t),k}_n -\eta^{(t),k} \widetilde{\nabla} \mathcal{L}_n(\bm{\omega}^{(t),k}_n)$, where $\eta^{(t),k}$ denotes the step-size/learning-rate, and $\mathcal{L}_n(\cdot)$ denotes the local loss function of the institution $n$, which quantifies the predictive accuracy of the ML model over its local dataset. Further, $\widetilde{\nabla} \mathcal{L}_n(\cdot)$ denotes the stochastic gradient obtained by sampling a random mini-batch of data from the local dataset $\mathcal{D}^\mathsf{Tr}_n$.

\vspace{-0mm}
\begin{table}
\caption{Major notations used in the paper.}
\label{tab:notation-table}
% \vspace{-2mm}
\scriptsize
\begin{tabularx}{0.49\textwidth}{| >{\centering\arraybackslash}m{15mm}
                                 | >{\centering\arraybackslash}X |}

  \hline
  \rowcolor{DarkBlue}\textbf{Notation} & \textbf{Description}\\
  \hline \hline
  $\mathcal{N}$ & Set of participating institutions \\
  \hline
  \rowcolor{LightBlue}$n$ & Sample institution index \\
  \hline
  $\mathcal{D}_n$ & Set of local datapoints of institution $n$ \\
  \hline
  \rowcolor{LightBlue}$\mathcal{D}_n^{\mathsf{Tr}}$ & Set of local train datapoints of institution $n$ \\
  \hline
  $\mathcal{D}_n^\mathsf{Va}$ & Set of local validation datapoints of institution $n$ \\
  \hline
  \rowcolor{LightBlue}$\mathcal{D}$ & Global dataset \\
  \hline
  $d$ & Sample datapoint consisting of features and class label \\
  \hline
  \rowcolor{LightBlue}$\bm{X}$ & Set of input features of an arbitrary datapoint\\
  \hline
  $\bm{y}$ & Class label of an arbitrary datapoint\\
  \hline
  % $\widehat{y}$ & Predicted class label \\
  % \hline
  \rowcolor{LightBlue}$\bm{x}_m$ & Input features for modality $m$ of a datapoint $d = (\bm{X}$, $\bm{y})$\\
  \hline
  $K$ & Number of local training rounds across institutions \\
  \hline
  \rowcolor{LightBlue}$k$ & Local training round index \\
  \hline
  $T$ & Number of global aggregation steps \\
  \hline
  \rowcolor{LightBlue}$t$ & Global aggregation step index \\
  \hline
  $\bm{\omega}_n^{(t),k}$ & Local ML model parameters for institution $n$ at global aggregation step $t$ and local training round $k$ \\
  \hline
  \rowcolor{LightBlue}$\bm{\omega}_{m,n}^{(t),k}$ & Institution $n$'s encoder parameters for modality $m$ at global aggregation $t$ and local round $k$ \\
  \hline
  $\bm{\varpi}_n^{(t),k}$ & Classifier parameters for institution $n$'s ML model \\
  \hline
  \rowcolor{LightBlue}$\widehat{\bm{\omega}}_{m}^{(t)}$ & Global encoder of modality $m$ at global aggregation step $t$ \\
  \hline
  $\widehat{\bm{\varpi}}_{C}^{(t)}$ & Global classifier of modality combination $C$ at global aggregation step $t$\\
  \hline
  \rowcolor{LightBlue}$\bm{\omega}^{(t)}$ & Global model parameters at global aggregation step $t$ \\
  \hline
  $\bm{\omega}_{\mathsf{obj}}$ &  Set of all the encoder and classifier parameters across the institutions\\
  \hline
  \rowcolor{LightBlue}
  $\bm{\omega}_{\mathsf{obj}}^{\star}$ &  Optimal set of ML model weights for all institutions\\
  \hline
  $\mathcal{B}_n^{(t),k}$ & Stochasticly selected mini-batch of data from institution $n$'s local dataset at global aggregation step $t$ and local training round $k$\\
  \hline
  \rowcolor{LightBlue}$\eta^{(t),k}$ & Learning rate at global aggregation $t$ and local round $k$ \\
  \hline
  $\widetilde{\nabla} \mathcal{L}_n$ & Stochastic gradient of institution $n$ \\
  \hline
  \rowcolor{LightBlue}$\mathcal{M}$ & Set of all available modalities across the institutions\\
  \hline
  % \rowcolor{LightBlue}$\mathcal{Y}$ & Set of all available class labels across the institutions\\
  % \hline
  $\mathcal{M}_n$ & Data modality combination held by institution $n$\\
  \hline
  \rowcolor{LightBlue}$P(\mathcal{M})$ & Set of all existing modality combinations in the network\\ 
  \hline
  $C$ & An arbitrary combination of modalities \\
  \hline
  \rowcolor{LightBlue}$\overline{\rho}_{C,n}^{(t)}$ & Normalized client weight for institution $n$ with modality combination $C$ at global aggregation step $t$\\
  \hline
  % $a_{m,n}$ & Binary indicator of whether institution $n$ has data of modality $m$\\
  % \hline
  % $b_{C,n}$ & Binary indicator of whether institution $n$ has data of modality combination $C$\\
  % \hline
  $\mathcal{L}_n^{\mathsf{Va}}$ & Validation loss of institution $n$\\
  \hline
  \rowcolor{LightBlue}$\mathcal{L}_n^{\mathsf{Tr}}$ & Training loss of institution $n$\\
  \hline
  $O_C^{(t)}$ & Overfitting for modality combination $C$ at global aggregation step $t$\\
  \hline
  \rowcolor{LightBlue}$G_C^{(t)}$ & Generalization for modality combination $C$ at global aggregation step $t$\\
  \hline
  % $\overline{\mathcal{L}}_C^{\mathsf{Va},(t)}$ & Average of the validation loss from institutions training on data with modality combination C\\
  % \hline
  % $\overline{\mathcal{L}}_C^{\mathsf{Tr},(t)}$ & Average of the training loss from institutions training on data with modality combination C\\
  % \hline
  $\Gamma_{m,n}^{(t)}$ & DOGR coefficent at global aggregation step $t$ for the encoder of modality $m$ at institution $n$\\
  \hline
  \rowcolor{LightBlue}
  $\Gamma_{\mathcal{M}_n,n}^{(t)}$ & DOGR coefficent at global aggregation step $t$ for the classifier of institution $n$\\
  \hline
  $\varphi_{n}^{(t)}$ & DOGR normalization coefficient for institution $n$ at global aggregation step $t$\\
  \hline
  \rowcolor{LightBlue}
  $\widetilde{\nabla}_{m} \mathcal{L}_n$ & Stochastic gradient of institution $n$'s ML model with respect to the modality $m$ encoder parameters\\
  \hline
  $\widetilde{\nabla}_{\bm{\varpi}} \mathcal{L}_n$ & Stochastic gradient of institution $n$'s ML model with respect to the classifier parameters \\
  \hline
  \rowcolor{LightBlue}
  $\Delta\bm{\omega}_{C, n}^{(t)}$ & Cumulative gradient of institution $n$'s local ML model with modality combination $C$ in global aggregation step $t$\\
  \hline
  $\Delta\bm{\omega}_{C,G}^{(t)}$ & Gradient of the global model with modality combination $C$ from global aggregation step $t-1$ to $t$ \\
  \hline
\end{tabularx}
\end{table}

Upon acquiring $\bm{\omega}^{(t),K}_n$ at each institution $n$, the institution transmits it to the server through upstream/uplink communications. Subsequently, the server aggregates the gathered ML models of institutions $\{\bm{\omega}^{(t),K}_n\}_{n\in\mathcal{N}}$ to form a new global model $\bm{\omega}^{(t+1)}$ via the following process:
\begin{equation} \label{equ:unimodalfedavg}
    \bm{\omega}^{(t+1)} =\sum_{n \in\mathcal{N}}{D_n \bm{\omega}_n^{(t),K}}\big/{D}.
\end{equation}
This global model is broadcast to institution, using which they synchronize their local models as mentioned above (i.e., $\bm\omega^{(t+1),0}_n~\leftarrow~\bm\omega^{(t+1)},~~\forall n$) and initiate the next round of local ML training and global aggregation.

\vspace{-3mm}
\subsection{Federated Learning over Unbalanced Modalities} \label{subsec:unbalanced-mmfl}

\subsubsection{Multi-modal FL under Unbalanced Modalities}
\begin{figure}[t]
\begin{center}
\centerline{\includegraphics[width=0.5\textwidth]{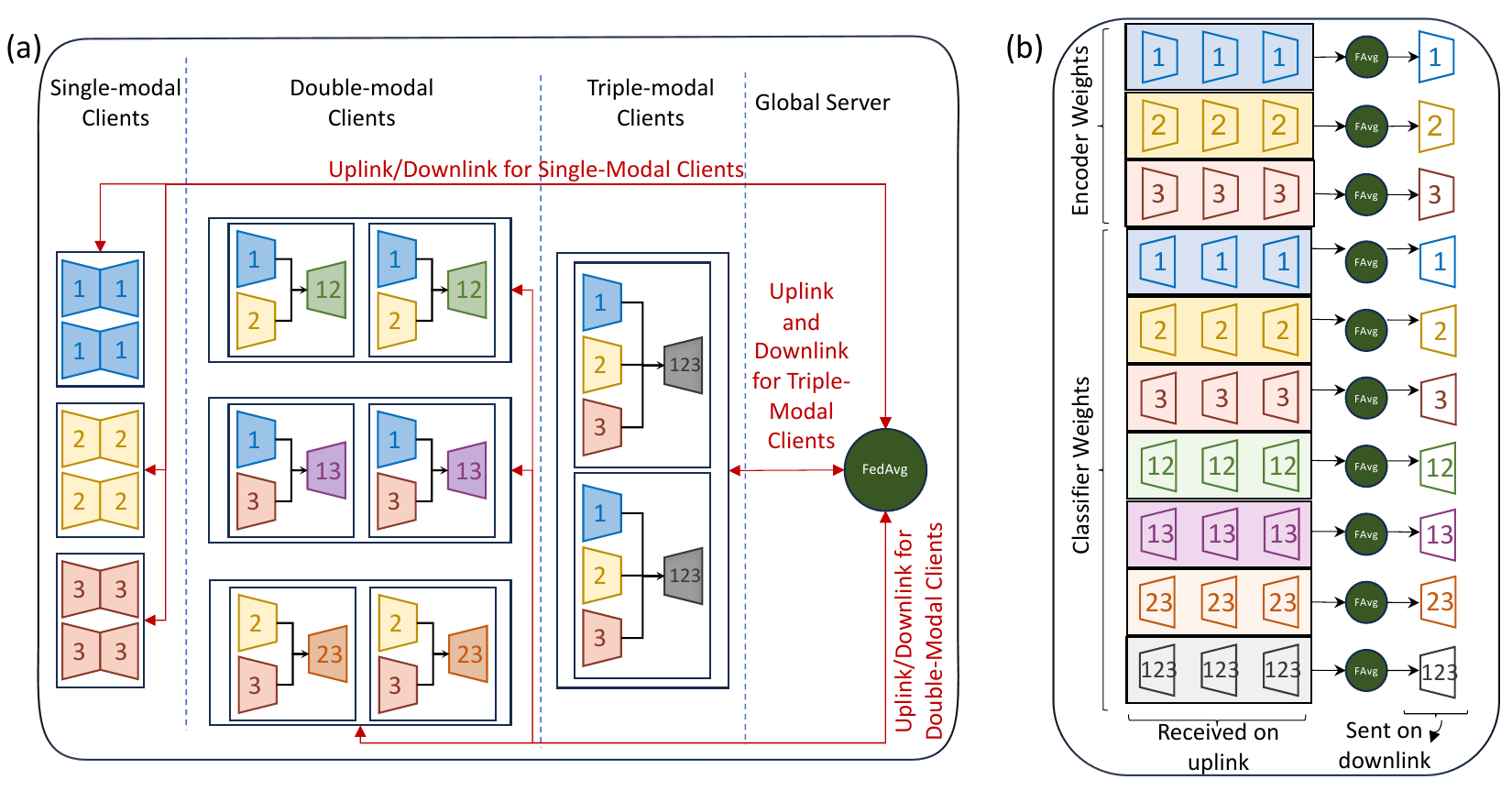}}
\vspace{-1.5mm}
\caption{(a) Our system model: each institution $n$ can have access to a subset of data modalities ($\mathcal{M}_n \subseteq \mathcal{M}$). (b) The server gathers encoders and classifiers and aggregates them to global models and sends them back to institutions.}
\label{fig:system-model}
\end{center}
\vspace{-4mm}
\end{figure}

In real-world scenarios, institutions may have access to various subsets of modalities. This phenomenon makes the conventional multi-modal FL, which is based on naïve extension of uni-modal FL to multi-modal settings \cite{mm-fl-gradblend, XIONG2022110, CREMONESI2023104338} impractical, as these works presume the existence of the same set of data modalities across all the institutions.
In particular, in a network of institutions with non-uniform number of data modalities, the number of encoders used across them will differ. Further, the size of the input feature fed to the classifier will also vary across institutions. These variabilities make the naïve ML training and aggregation methods of \cite{mm-fl-gradblend, XIONG2022110, CREMONESI2023104338} ineffective.
Consequently, we propose a generalized distributed encoder-classifier architecture, which increases the flexibility of model aggregations at the server by aggregating (i) each modality's encoder and (ii) classifiers of different existing modality combinations across the institutions separately. 
 A schematic of our proposed system and model aggregation strategy is depicted in Fig. \ref{fig:system-model}(a) and \ref{fig:system-model}(b), respectively.
 
Let $\mathcal{M}$ and $\mathcal{Y}$ denote the sets encompassing all data modalities within our network distributed across the institutions, and the data labels (i.e., cancer stages in our scenario), respectively. Moreover, let $\mathcal{M}_n \subseteq \mathcal{M}$ denote the subset of modalities held by institution $n$, which are present in its local dataset $\mathcal{D}_n$. Considering each institution $n$, each data point $d \triangleq (\bm{X}, \bm{y}) \in \mathcal{D}_n$ is associated with a feature vector $\bm{X}$ and a label $\bm{y}$, where $\bm{X} = \{\mathbf{x}_m\}_{m\in \mathcal{M}_n}$ contains the set of features corresponding to each modality $m$ possessed by the institution. Additionally, $\bm{y}\in \mathcal{Y}$ is represented in a one-hot encoded format, capturing the cancer stage.
As will be formalized next, to handle institutions with an uneven distribution of modalities, we will put a greater emphasis on improving the local training of encoders. This approach diverges from the recent methodologies in multi-modal FL \cite{XIONG2022110}, where the emphasis lies on aggregating the classification layers -- and co-attention layers -- which adapt well to scenarios where institutions have access to all the modalities. This paradigm shift allows our approach to extend the distributed ML training framework of FL for each modality to all institutions possessing data pertaining to that specific modality. In particular, in our approach, the encoder used for each modality would benefit from the commonality of data across the institutions possessing that modality. This, in turn, eliminates the restricting assumption that institutions must possess all data modalities.

\begin{figure}[t]
\begin{center}
\centerline{\includegraphics[width=\columnwidth]{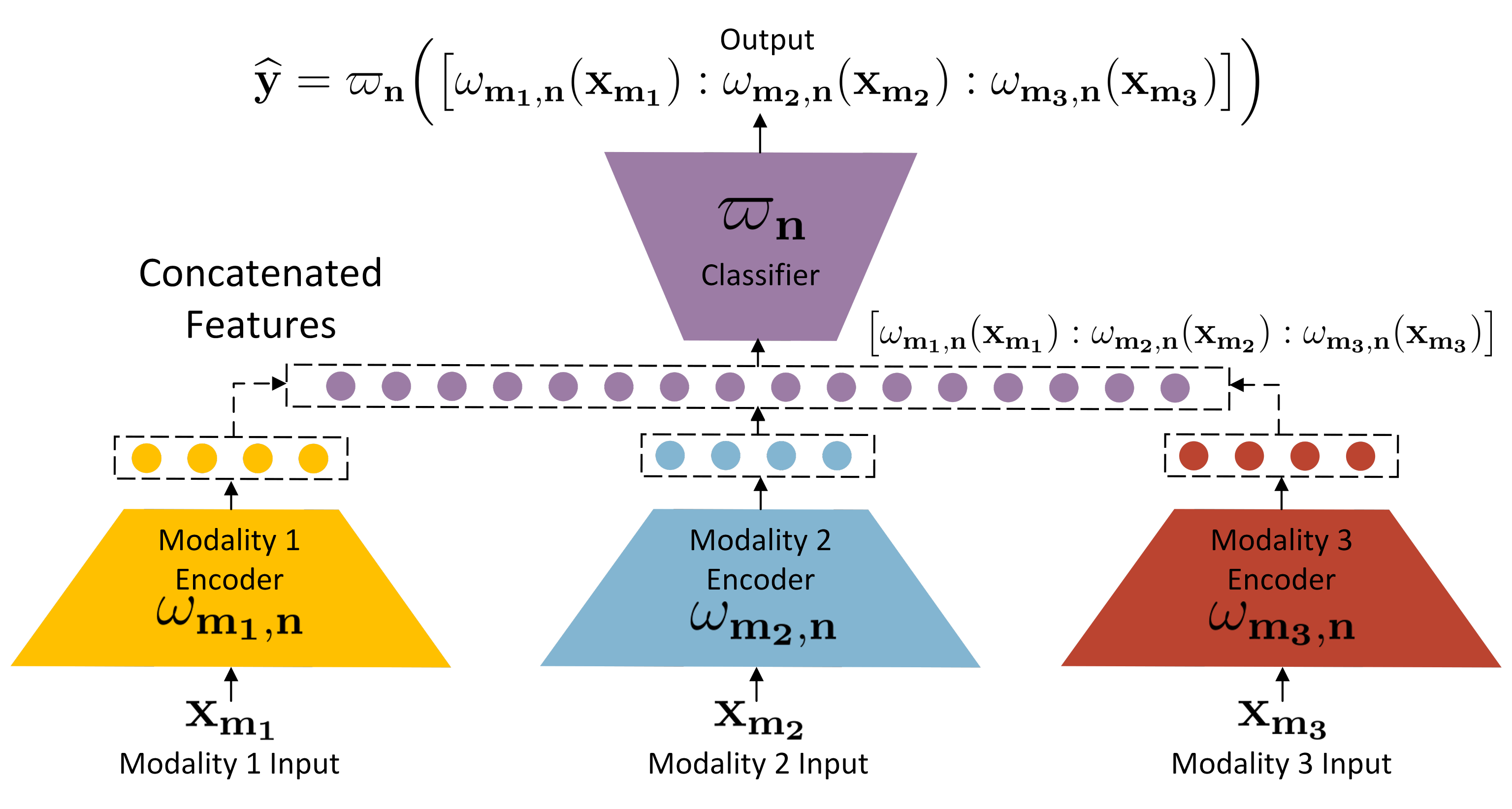}}
\vskip -0.12in
\caption{A schematic of the encoder-classifier structure at an arbitrary institution $n$ with three data modalities.}
\label{fig:mm-ed-struct}
\end{center}
\vskip -0.2in
\end{figure}

Considering encoder parameters/models $\{\bm{\omega}_{m,n}\}_{m\in\mathcal{M}_n}$ and the classifier parameters $\bm{\varpi}_{n}$ at institution $n$ (see Fig. \ref{fig:mm-ed-struct}), the ultimate predicted label $\widehat{y}$ for a datapoint $d\in\mathcal{D}_n$ is given by
\begin{equation*}
\label{equ:yhat-prediction}
   \hspace{-2mm} \widehat{y}  \hspace{-0.3mm}= \hspace{-0.3mm} \bm{\varpi}_n \hspace{-0.3mm} \Big( \hspace{-0.6mm}\big[ \hspace{-0.3mm}\bm{\omega}_{m_1,n} \hspace{-0.3mm}( \hspace{-0.3mm}\mathbf{x}_{m_1} \hspace{-0.3mm})  \hspace{-0.3mm}: \hspace{-0.3mm} \bm{\omega}_{m_2,n} \hspace{-0.3mm}( \hspace{-0.3mm}\mathbf{x}_{m_2} \hspace{-0.3mm})  \hspace{-0.3mm}:   \hspace{-0.3mm}\cdots  \hspace{-0.3mm}:  \hspace{-0.3mm}\bm{\omega}_{|\mathcal{M}_n|,n} \hspace{-0.3mm}( \hspace{-0.3mm}\mathbf{x}_{m_{|\mathcal{M}_n|}} \hspace{-0.3mm}) \hspace{-0.3mm}\big] \hspace{-0.6mm}\Big) \hspace{-0.3mm},\hspace{-2mm}
\end{equation*}
where $\bm{\omega}_{m_i, n}(\mathbf{x}_{m_i})$, for each $i\in\{1,\cdots,|\mathcal{M}_n|\}$, represent the encoder output upon being fed with data feature of modality $m_i$. Also, the function $\bm{\varpi}_n(\cdot)$ encompasses the description of the classifier, which takes as its input the concatenated/fused feature of all the modalities available at the institution and produces a predicted label as its output.

\subsubsection{Machine Learning Task Formulation}

We consider the loss function at institution $n$ for an arbitrary local model parameter realization $\bm{\omega}_n$ as $\mathcal{L}_n(\bm{\omega}_n; \mathcal{D}_n) = \frac{1}{|\mathcal{D}_n|} \sum_{d \in \mathcal{D}_n} \ell(\bm{\omega}_n; d)$ where $\ell(\bm{\omega}_n; d)$ is the ML model loss function evaluated on datapoint $d$. In this work, we use cross-entropy to measure the loss \cite{good1952rational}.
Also, we use $\bm{\omega}_{n} \triangleq \left[ \{\bm{\omega}_{m,n}\}_{m \in \mathcal{M}_n} : \bm{\varpi}_{n}\right]$ to denote the concatenation of the encoders model parameters in conjunction with the classifier for institution $n$.

Since in our system of interest, institutions have different combinations of modalities, to capture the present combinations of modalities in the system, we introduce a set\footnote{Note that  $\mathcal{P}(\mathcal{M})$ is a set and thus only has non-identical elements.} 
\begin{equation}
    \mathcal{P}(\mathcal{M}) \triangleq \{\cup_{n \in \mathcal{N}} \mathcal{M}_n\}.
\end{equation}
If all the institutions have the same modality ($\mathcal{M}_n=\mathcal{M}, \forall n$), then  $ \mathcal{P}(\mathcal{M})$=$\{\mathcal{M}\}$.
Subsequently, since institutions with the same combinations of modalities use the same architecture of classifiers, we denote $\widehat{\bm{\varpi}}_C$ as the common architecture of the classifier used in the system, for all institutions $n$ with modality combination $C\in\mathcal{P}(\mathcal{M})$, for which $|\widehat{\bm{\varpi}}_C|=|\bm{\varpi}_{n}|$, $\forall n$, where $|.|$ captures the number of classifier parameters.\footnote{For example, if the system has one institution with modality $m_1$ (i.e., $\mathcal{M}_1=\{m_1\}$), one institution with modality $m_2$, one institution with modality $m_3$, one institution with modality $m_1,m_2$, and one institution with modality $m_1, m_2, m_3$, then,  $\mathcal{P}(\mathcal{M})=\{ \{m_1\},\{m_2\},\{m_3\},\{m_1,m_2\}, \{m_1,m_2,m_3\} \}$. In this case, $C$ can be any of the five sets that constitute $\mathcal{P}(\mathcal{M})$.}

Subsequently, we can define the global loss function as
\begin{gather} \label{equ:objective-function}
    \mathcal{L}(\bm{\omega}_{\mathsf{obj}}) \triangleq  \sum_{n \in \mathcal{N}} \frac{|\mathcal{D}_n|}{|\mathcal{D}|}\mathcal{L}_n(\bm{\omega}_n),
\end{gather} 
where $\bm{\omega}_{\mathsf{obj}}$ encompasses all the encoder and classifier parameters across the institutions (i.e., $\bm{\omega}_{\mathsf{obj}}= \{\bm{\omega}_{n}\}_{n\in\mathcal{N}}$).

Therefore, our goal is to find a set of global model parameters, consisting of a set of unified model parameters for encoders used for different modalities and a set of unified classifiers used for existing modality combinations in the network. Mathematically, we are interested in obtaining $\widehat{\bm{\omega}}_{\mathsf{obj}}^\star$, which is given by 
\begin{equation} \label{equ:min-omega}
  \hspace{-3mm} \widehat{\bm{\omega}}_{\mathsf{obj}}^\star= \big\{\{\widehat{\bm{\omega}}^\star_{m}\}_{m \in \mathcal{M}}, \{\widehat{\bm{\varpi}}_{C}^\star\}_{C\in\mathcal{P}(\mathcal{M})}\big\} \hspace{-0.5mm}\triangleq \hspace{-0.1mm} \argmin_{\widehat{\bm{\omega}}_\mathsf{obj}}\hspace{-0.1mm} \mathcal{L}(\widehat{\bm{\omega}}_{\mathsf{obj}}), \hspace{-2mm} 
\end{equation}
where $\widehat{\bm{\omega}}_{\mathsf{obj}}$ is defined similarly to $\bm{\omega}_{\mathsf{obj}}$ in~\eqref{equ:objective-function} upon deployment of unified encoder and classifier parameters across the institutions possessing the same modality combinations; mathematically
$
\widehat{\bm{\omega}}_{\mathsf{obj}} = \{\bm{\omega}_{n}\}_{n\in\mathcal{N}}=\{\left[ \{\bm{\omega}_{m,n}\}_{m \in \mathcal{M}_n} : \bm{\varpi}_{n}\right]\}_{n\in\mathcal{N}}$ if $ \bm{\omega}_{m,n} = \widehat{\bm{\omega}}_{m}~ \forall n$ and  $\bm{\varpi_{n}} = \widehat{\bm{\varpi}}_{C}$ if institution $n$ possesses modality combination $C$. In particular, $\widehat{\bm{\omega}}_{\mathsf{obj}}^\star$ is the \textit{set} of all \textit{global} optimal parameters of all encoders and classifiers present in the network, where $\widehat{\bm{\omega}}_{m}$ refers to the unified encoder parameter used for modality $m$ across all the institutions, and $\widehat{\bm{\varpi}}_{C}$ refers to the unified parameters of the classifiers used in institutions possessing the modality combination $C$ in the network. In the following, we describe the iterative procedure of local and global model aggregations, aiming to learn the optimal values of these parameters, i.e., $\{\widehat{\bm{\omega}}^\star_{m}\}_{m \in \mathcal{M}}, \{\widehat{\bm{\varpi}}_{C}^\star\}_{C\in\mathcal{P}(\mathcal{M})}$.\footnote{It is noteworthy to mention that upon having the \textit{same set of data modalities} across the institutions, $\mathcal{P}(\mathcal{M}) = \{\mathcal{M}\}$, $\widehat{\bm{\omega}}_{\mathsf{obj}}$ would contain the global encoder and classifier parameters that will be deployed uniformly across all the institutions upon the conclusion of distributed ML model training.}

\subsubsection{Local ML Model Training}
At each institution $n$, the local model training can be described via a sequence of mini-batch SGD iterations. In particular, given a local model at an institution at SGD iteration $k$, at global model aggregation step $t$ (i.e., $\bm{\omega}_n^{(t),k}$), the next local model is obtained as follows:
\vspace{-3mm}

{ \small
\begin{equation} \label{equ:local-update}
    \bm{\omega}^{(t), k+1}_n = \bm{\omega}^{(t), k}_n - \eta^{(t), k} \widetilde{\nabla}\mathcal{L}_n(\bm{\omega}^{(t), k}_n; \mathcal{B}_n^{(t),k}),~0 \leq  k\leq K-1
\end{equation}
}
\vspace{-3mm}

\noindent where
\begin{gather} \label{equ:local-batch-loss}
    \widetilde{\nabla}\mathcal{L}_n(\bm{\omega}^{(t), k}_n; \mathcal{B}_n^{(t), k}) \triangleq \frac{1}{|\mathcal{B}_n^{(t), k}|} \sum_{d \in \mathcal{B}_n^{(t), k}} \nabla \ell(\bm{\omega}^{(t), k}_n; d),
\end{gather}
and
$
    \bm{\omega}_n^{(t),k} = \big\{\{\bm{\omega}_{m,n}^{(t),k}\}_{m \in \mathcal{M}_n}: \bm{\varpi}^{(t),k}_n\big\}.
$
In \eqref{equ:local-batch-loss}, $\mathcal{B}_n^{(t), k}$ denotes a mini-batch of data sampled uniformly at random from $\mathcal{D}_n^{\mathsf{Tr}}$ and $\eta^{(t), k}$ is the learning-rate. Assuming that institutions conduct $K$ SGD iterations before transmitting their models to the server, we let $ \bm{\omega}_n^{(t),K} = \big\{\{\bm{\omega}_{m,n}^{(t),K}\}_{m \in \mathcal{M}_n}: \bm{\varpi}^{(t),K}_n\big\}$ denote the final local encoder and classifier parameters at institution $n$.

\subsubsection{Local Model Transfer and Global Model Aggregations}

At the end of each local training round, after receiving the parameters of the encoders and classifiers from the institutions, the server then aggregates them as follows to obtain a unique encoder parameter set for each modality and a unique classifier parameter set for each of the existing combinations of modalities in the system:
\begin{align} \label{equ:aggregation-encoder}
 \widehat{\bm{\omega}}_{m}^{(t+1)} &= \frac{\sum_{n\in\mathcal{N}} a_{m,n} |\mathcal{D}_n| \bm{\omega}_{m,n}^{(t),K}}{\sum_{n\in\mathcal{N}} a_{m,n} |\mathcal{D}_n|}, \forall m \in \mathcal{M},  \\
 \label{equ:aggregation-classifier}
 \widehat{\bm{\varpi}}_{C}^{(t+1)} &= \frac{\sum_{n\in\mathcal{N}} b_{C,n} |\mathcal{D}_n| \bm{\varpi}_{n}^{(t),K}}{\sum_{n\in\mathcal{N}} b_{C,n} |\mathcal{D}_n|}, \forall C \in \mathcal{P}(\mathcal{M}),
\end{align}
where $a_{m,n}$, $\forall n,m$ is a binary indicator which takes the value of $1$ if institution $n$ has access to modality $m$; otherwise $a_{m,n} = 0$. 
Similarly, $b_{C,n}$ is a binary indicator that takes the value of $1$ when the modalities at institution $n$ match the combination $C$ (i.e., $\mathcal{M}_n = C$); otherwise $b_{C,n}=0$. After conducting the model aggregation step, the new sets of encoders and classifiers are sent back in the downstream link to the institutions, which are then used to synchronize the local models. In particular, each institution $n$ with modality combination $C$ will conduct ${\small \bm{\omega}_{m,n}^{(t+1), 0} \leftarrow \widehat{\bm{\omega}}_{m}^{(t+1)}}$ and  ${ \small \bm{\varpi}_{n}^{(t+1), 0} \leftarrow \widehat{\bm{\varpi}}_{C}^{(t+1)}}$ (i.e., $ \bm{\omega}_n^{(t+1),0} = \big[\{\widehat{\bm{\omega}}_{m}^{(t+1)}\}_{m\in\mathcal{M}_n} : \widehat{\bm{\varpi}}_{C}^{(t+1)}\big]$). This synchronization will be followed by SGD iterations described in \eqref{equ:local-update} and then the next global aggregation step. This procedure continues until reaching a desired accuracy.

\begin{figure*}[t]
\vspace{-3mm}
    \centering
    \includegraphics[width=0.329\textwidth]{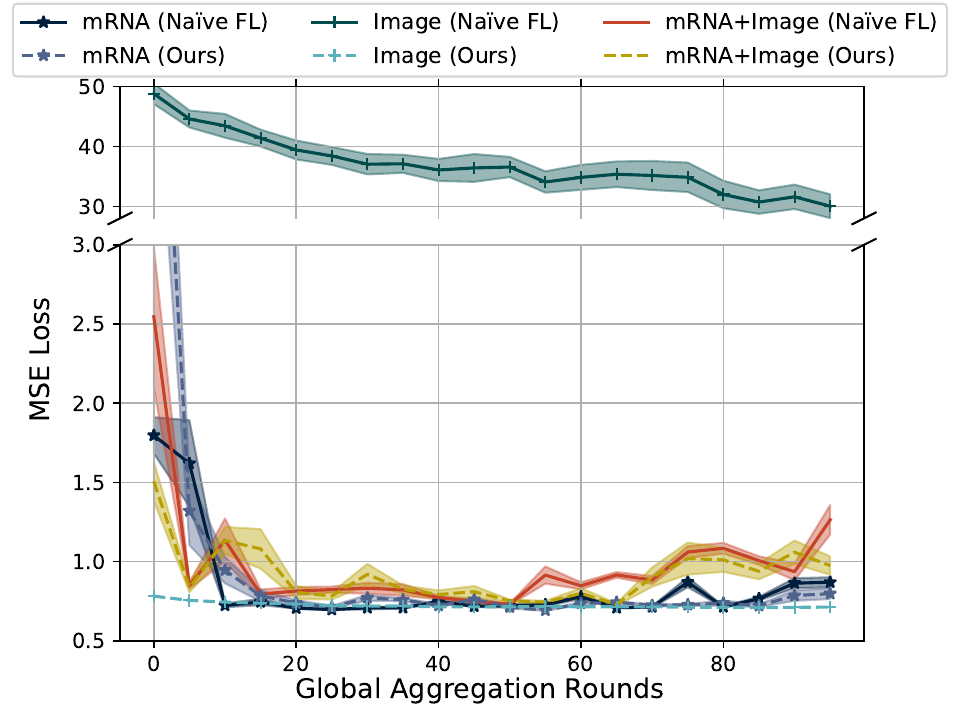} \hfill
    \includegraphics[width=0.329\textwidth]{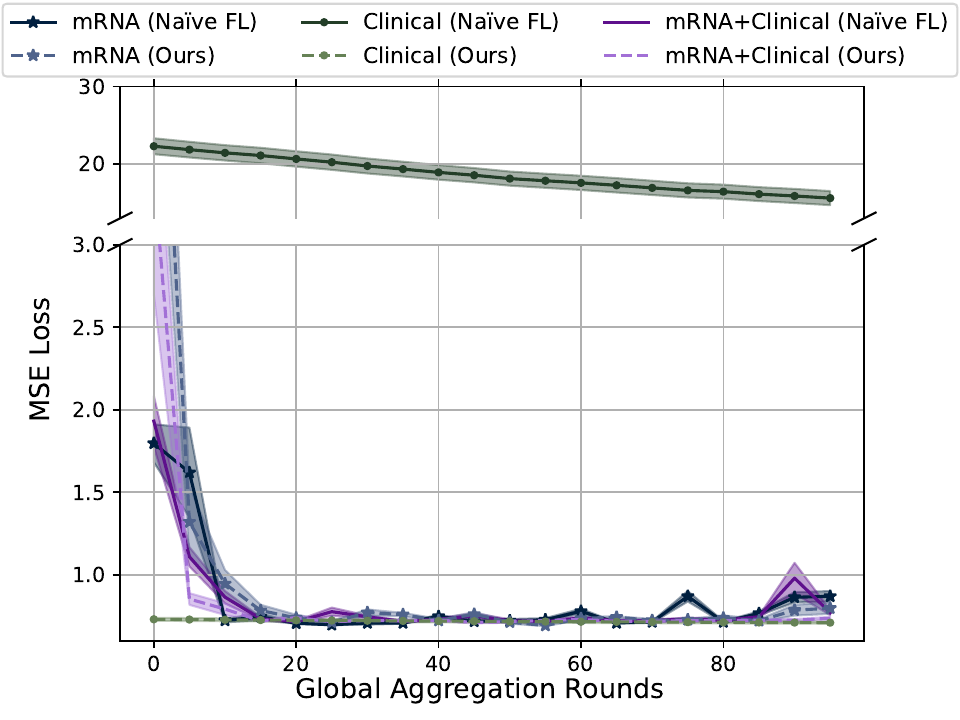} \hfill
    \includegraphics[width=0.329\textwidth]{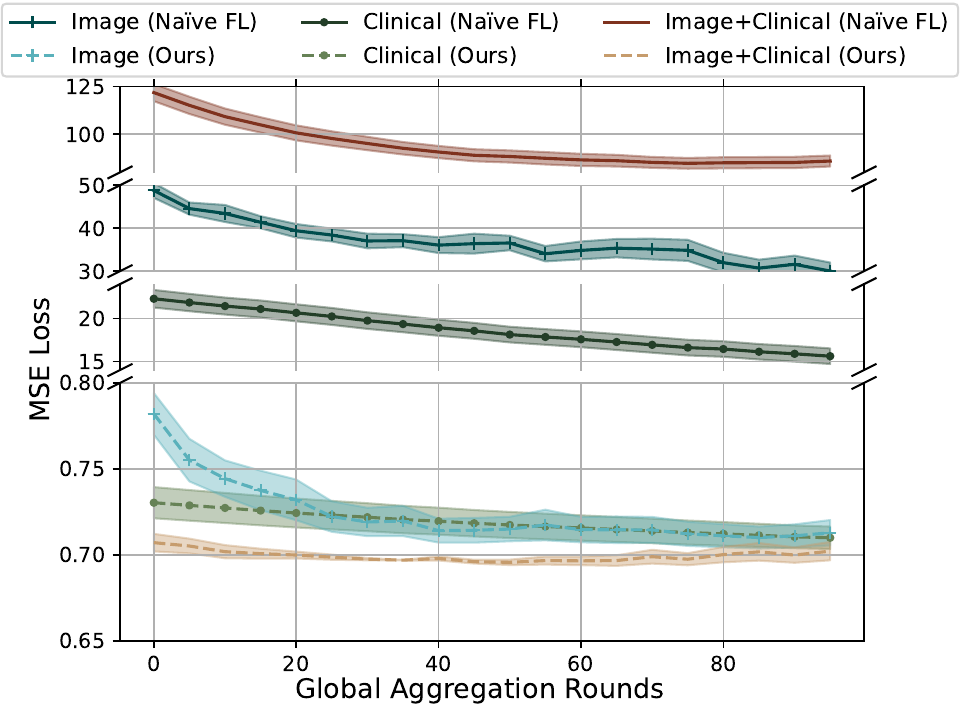}
    \vspace{-6mm}
    \caption{
 Loss plots of the global models based on the modalities possessed across institutions in the first 100 global aggregation steps. 
 Plots consider various modality combinations: (a)  image and mRNA modalities, (b) clinical and mRNA, and (c) image and clinical. 
 The shaded areas represent the noise (20\% standard deviation) for each combination of modalities tested. A key observation is the occasional degradation of the performance of the multi-modal ML models compared to their single-modal counterparts. This observation is consistent with the claims in \cite{mm-fl-gradblend} which emphasises on the necessity of syncing the convergence rates of the modalities in a multi-modal FL scenario. Further, our method leads to the balance of convergence rates across the modalities. 
  }
    \label{fig:not-converging}
    \vspace{-5.5mm}
\end{figure*}

\vspace{-2mm}
\subsection{Key Challenges and Design Criteria} \label{subsec:key-challenges}

Our proposed system model contains nuances in decoupling the aggregation of encoders and classifiers, which is necessary in handling institutions with an unbalanced number of modalities. However, achieving a high prediction performance across this system faces a major obstacle caused by \textit{straggling modalities}. We demonstrate this phenomenon in Fig. \ref{fig:not-converging}, in which we conduct ML model training according to Sec. \ref{subsec:unbalanced-mmfl}, which implies training all institutions with unified learning-rates (labelled as \textit{Naïve FL} in the plots). We consider a dataset of three modalities for cancer staging, consisting of mRNA sequences, histopathological images, and clinical information and presume the existence of $9$ institutions, where $3$ have access to only one modality of data, $3$ to two modalities of data, and $3$ to all modalities of data. The details of the dataset used for patients are given in Sec. \ref{subsec:data}. As can be seen in the figure, the performance of some 
modalities is notably better than the others, as the losses are decreasing faster for those modalities  but the same does not apply to  others. Furthermore, during the local training phase, the ML model of some modality combinations has converged notably faster than the rest. Also, some modalities struggle to improve their loss and accuracy during the ML model training. 
We next aim to propose a framework to address these issues, the result of which is labelled as \textit{ours} in the plots in Fig.~\ref{fig:not-converging}. It can be observed from the curves that our method balances the convergence rates of various modality combinations, leading to a notable performance gain as verified later in Sec. \ref{sec:experiments}.

\vspace{-1.5mm}
\section{Distributed Gradient Blending and Proximity-Aware Client Weighting Approach}
\label{sec:resolution}
\noindent In the following, we first provide an overview of our methodology to address the above-described issues of multi-modal FL (Sec.~\ref{sub:MMFLIssuRes}). We then describe the two key components of our method which consists of distributed gradient blending (Sec.~\ref{sub:DGB}) and proximity-aware client weighting  (Sec.~\ref{sub:PCW}).

\vspace{-3.5mm}
\subsection{Tackling Multi-Modal FL with Unbalanced Modalities}\label{sub:MMFLIssuRes}
Considering the discussion in Sec. \ref{subsec:key-challenges},  we aim to propose a methodology for multi-modal FL that alters the local gradient computations to solve the problem of different convergence rates of various modalities. In a nutshell, our method consists of two components: (i) distributed gradient blending (DGB), which is a methodology to evaluate the contribution of various modalities to the global model training without transferring the data across the network, which is then used to control the convergence rates of modalities so as to accelerate the FL model convergence; and (ii) proximity-aware client weighting (PCW), which is a method for tackling the impact of non-iid data and biased local datasets at the institutions via evaluating the quality of data possessed by institutions -- without data transfer across the network -- via a similarity metric. This metric is then used to quantify the extent of reliance on the performance of each local model on calculating the DGB parameters. We next provide a high-level description of these two key components of our method, and mathematically formalize them later in Sec.~\ref{sub:DGB} and~\ref{sub:PCW}:

\begin{itemize}[leftmargin=4mm]
    \item \textit{Distributed Gradient Blending (DGB):} To remove the barrier of non-uniform convergence speeds across various data modalities, DGB weighs the gradients that are being back-propagated through each encoder in an institution's model  to unify the convergence rates of modalities. DGB achieves this through acquiring a set of train and validation losses from each institution at the end of each local ML training round, which is in turn used to calculate two metrics of \textit{overfitting} and \textit{generalization}. These
    two metrics are then used to calculate the distributed overfitting
    to generalization ratio (DOGR), formalized in Sec.~\ref{sub:DGB}. 
    % The DOGR will be used to weigh
    % the gradients during local model training in order to control the
    % speed of convergence for each modality, 
    % since the converging 
    % set of weights for a specific modality might result in a sub-optimal
    % performance for another set. 
    The DOGR will measure the change in 
    overfitting and generalization ability of the present modalities 
    across the available institutions to quantify the average performance improvement for each modality combination across institutions in the recent global aggregation step.
    DOGR is then integrated to
    obtain the aforementioned weights
    for the gradients, which can accelerate or slacken the ML
    training rate of different modalities at various institutions.
    \item \textit{Proximity-Aware Client Weighting (PCW):} PCW mitigates the performance degradation of DGB caused by the heterogeneity of data across institutions, which leads to the bias of the locally trained ML models, especially as the length of local ML model training (i.e., $K$) grows. To overcome the effect of non-iid data distribution across institutions on the losses acquired from them during the calculation of DOGR weights, PCW aims to quantify the extent of reliance on the local losses of various institutions -- used in computing the DOGR parameters -- according to the quality of data held by institutions. To this end, PCW
    weighs the received local losses of institutions at the server based on the trajectory of both local and global models.
    Specifically, PCW promotes more reliance on the local losses of institutions whose local ML parameters' trajectory exhibits a closer alignment with that of the global model. This signifies that institutions with superior data quality (i.e., diverse local datasets that better encapsulate the global data distribution, and thus do not induce bias onto the locally trained ML model), will have a more pronounced influence on the calculation of DGB weights. By mitigating the impact of local ML model bias, PCW makes DGB robust against the increase in the number of local SGD iterations $K$ in each global aggregation (i.e., reductions in uplink transmission frequency). The mathematical explanation of PCW is discussed in Sec.~\ref{sub:PCW}.
\end{itemize}
We next describe the working mechanism of DGB and PCW.

\vspace{-2mm}
\subsection{Distributed Gradient Blending (DGB)}
\label{sub:DGB}

To realize DGB, we first adapt and extend the notions of \textit{generalization} and \textit{overfitting}, defined in \cite{gradient-blending}, to the federated setting. These notions provide insights into the effect of local ML training with a given unified step size across modalities on the performance of the local ML models. In DGB, we use both the validation loss and training loss across institutions to compute the local gradient back-propagation weights for the upcoming local training rounds. In particular, let $\mathcal{L}_n^{\mathsf{Va}}$ denote the validation loss for institution $n$'s model at global aggregation step $t$ at SGD iteration $k$, which is  
\begin{align} \label{equ:validation-loss}
    \mathcal{L}_n^{\mathsf{Va}}(\bm{\omega}_n^{(t), k}) \triangleq &  \mathcal{L}_n(\bm{\omega}_n^{(t), k}; \mathcal{D}^{\mathsf{Va}}_n)  \nonumber \\
    = & \frac{1}{|\mathcal{D}^{\mathsf{Va}}_n|} \sum_{d \in \mathcal{D}_n^{\mathsf{Va}}} \ell(\bm{\omega}_n^{(t), k}; d), \bm{\omega}_n^{(t), k} \in \mathbb{R}^{|\bm{\omega}_n|}.
\end{align}
We similarly define the training loss for institution $n$'s model at SGD iteration $k$ of global aggregation step $t$ as follows:
\begin{align} \label{equ:train-loss}
    \mathcal{L}_n^{\mathsf{Tr}}(\bm{\omega}_n^{(t), k}) \triangleq &  \mathcal{L}_n(\bm{\omega}_n^{(t), k}; \mathcal{D}^{\mathsf{Tr}}_n)  \nonumber \\
    = & \frac{1}{|\mathcal{D}^{\mathsf{Tr}}_n|} \sum_{d \in \mathcal{D}_n^{\mathsf{Tr}}} \ell(\bm{\omega}_n^{(t), k}; d), \bm{\omega}_n^{(t), k} \in \mathbb{R}^{|\bm{\omega}_n|}.
\end{align}

We next derive two metrics called \textit{overfitting} $O_C^{(t)}$ and \textit{generalization} $G_C^{(t)}$ for each modality combination $C$ across the system, where $C\in \mathcal{P}(\mathcal{M})$, defined as\footnote{In the following, we assume that at least one institution with single-modal data for each modality $m$ (i.e., institution $n$ with $\mathcal{M}_n=\{m\}$) exists in the system. In other words, single-modal institutions cover all the modalities of data present in the system.}
\begin{align} 
    O_C^{(t)} &= \overline{\mathcal{L}}_C^{\mathsf{Va},(t)} - \overline{\mathcal{L}}_C^{\mathsf{Tr},(t)},\label{equ:o-def}\\ 
    G_C^{(t)} &= \overline{\mathcal{L}}_C^{\mathsf{Va},(t)}, \label{equ:g-def}
\end{align}
where {\small $\overline{\mathcal{L}}_C^{\mathsf{Va},{(t)}}$} and  {\small$\overline{\mathcal{L}}_C^{\mathsf{Tr},{(t)}}$} are the \textit{average/aggregation of the train and validation loss from institutions} training on data with modality combination $C$ (i.e., from institutions $n$ such that $\mathcal{M}_n=C$), obtained based on \eqref{equ:validation-loss} and \eqref{equ:train-loss} as later formalized through PCW in Sec. \ref{sub:PCW}. It is worth mentioning that in the previous approaches toward gradient blending, e.g., OGR metric in \cite{gradient-blending},  these metrics were calculated locally for each institution, where the losses in \eqref{equ:validation-loss} and \eqref{equ:train-loss} were directly used in \eqref{equ:o-def} and \eqref{equ:g-def} without averaging.\footnote{This 
will make the methodology prone to the bias of local datasets in our setting. In particular, biased and scarce local data sets at various institutions, which are common in FL and medical settings, will render localized computation of these parameters followed by their local adaptation for tuning the learning-rates of various modalities ineffective. 
For example, one modality may contribute to the training highly in one institution and also have a higher loss, making the trained model sensitive to the features of that specific modality. However, on a global scale, that modality may not have as high contribution and may also possess a small loss value when considered all the institutions.}

After obtaining $O_C^{(t)}$ and $G_C^{(t)}$ for all $C\in \mathcal{P}(\mathcal{M})$ at the server and broadcasting them across the institutions, at each global aggregation step $t$, these metrics are used in defining another metric, called Distributed Overfitting to Generalization Ratio (DOGR) for each institution $n$. 
In particular, each institution $n$ only uses a subset of $O_C^{(t)}$ and $G_C^{(t)}$ received from the server: only the values corresponding to the local encoders 
(i.e., $O_m^{(t)}$ for all $m\in\mathcal{M}_n$ obtained via $O_m^{(t)}=O_C^{(t)}$ when $C=\{m\}$; $G_m^{(t)}$ for all $m\in\mathcal{M}_n$ obtained via $G_m^{(t)}=G_C^{(t)}$ when $C=\{m\}$) and the overall combination held by the institution (i.e., $O_{\mathcal{M}_n}^{(t)}$ obtained via $O_{\mathcal{M}_n}^{(t)}=O_C^{(t)}$ when $C=\mathcal{M}_n$; $G_{\mathcal{M}_n}^{(t)}$ obtained via $G_{\mathcal{M}_n}^{(t)}=G_C^{(t)}$ when $C=\mathcal{M}_n$). Subsequently, the DOGR at each institution $n$ is computed for each modality $m$ (i.e., corresponding to each encoder) and the overall modality set held by the institution $\mathcal{M}_n$ (i.e., corresponding to the classifier) as follows:
\begin{equation} \label{equ:original-OGR2}
\begin{aligned}
    &\Gamma_{m,n}^{(t)} =  \frac{1}{\varphi^{(t)}_n} \times \frac{\left|\Delta G_m^{(t)}\right|^{2}}{\left(\Delta O_m^{(t)}\right)^2}, ~~m\in\mathcal{M}_n, 
    \\
    &\Gamma_{\mathcal{M}_n,n}^{(t)} =  \frac{1}{\varphi^{(t)}_n} \times \frac{\left|\Delta G_{\mathcal{M}_n}^{(t)}\right|^{2}}{\left(\Delta O_{\mathcal{M}_n}^{(t)}\right)^2}.
    \end{aligned}
\end{equation}
where {\small$\varphi_n^{(t)} = \frac{|\Delta G_{\mathcal{M}_n}|^{2}}{2(\Delta O_{\mathcal{M}_n})^2}+ \sum_{m \in \mathcal{M}_n} \frac{|\Delta G_m|^{2}}{2(\Delta O_m)^2}$} is a normalization coefficient obtained over the encoder modalities and the classifier combination existing in institution $n$. Also, {\small $\Delta O_m^{(t)} = O_m^{(t)} - O_m^{(t-1)}$} and {\small $\Delta G_m^{(t)} = G_m^{(t)} - G_m^{(t-1)}$}, where $m\in \mathcal{M}_n$, represent the variations in overfitting  and generalization, respectively (similarly $\Delta O_{\mathcal{M}_n}^{(t)}=O_{\mathcal{M}_n}^{(t)}-O_{\mathcal{M}_n}^{(t-1)}$ and $\Delta G_{\mathcal{M}_n}^{(t)}=G_{\mathcal{M}_n}^{(t)}-G_{\mathcal{M}_n}^{(t-1)}$). 
In particular, {\small $\Delta O_m^{(t)}$} captures the change in the overfitting during the $t$-th global aggregation step for modality $m$ under the used step size. Also,
{\small $\Delta G_m^{(t)}$} captures the change in the validation loss during the latest global aggregation step for modality $m$, which measures the overall performance variation of the respective modality. 
Consequently, $\Gamma_{m,n}^{(t)}$ is a metric that takes smaller values when the generalization of the modality encoder $m$ present at institution $n$ is decreasing or the model is becoming overfit during the global round and takes larger values  otherwise.
Thus, we can use $\Gamma_{m,n}^{(t)}$ to accelerate and slacken the learning-rate of different modality encoders present in institution $n$ according to their exhibited performance by increasing and dampening the unified SGD step-size during each local training round as discussed below.
The same discussion holds in the case of $\Delta O_{\mathcal{M}_n}^{(t)}$, $\Delta G_{\mathcal{M}_n}^{(t)}$, and $\Gamma_{\mathcal{M}_n, n}^{(t)}$ which are used to increase/decrease the SGD  step size in the classifier of each institution $n$.

We incorporate DOGR in the learning-rate of the local SGD iterations at each institution $n$. In particular, we modify the local SGD iterations given by \eqref{equ:local-update} to the following update rule, which accelerates/slackens the learning-rate across different modality subsets of various institutions ($0\leq k\leq K-1$):
\vspace{-3mm}

{\small
\begin{equation}\label{eq:update11}
  \hspace{-3mm}  \bm{\omega}^{(t), k+1}_{m,n} \hspace{-.5mm}=\hspace{-.5mm} \bm{\omega}^{(t), k}_{m,n} \hspace{-.5mm}-\hspace{-.5mm} \eta^{(t),k} \Gamma_{m,n}^{(t)} \widetilde{\nabla}_{m} \mathcal{L}_n(\bm{\omega}^{(t), k}_n; \mathcal{B}_n^{(t),k}),\hspace{-.5mm} \forall m \hspace{-.5mm} \in  \hspace{-.7mm}\mathcal{M}_n,\hspace{-2.5mm}
\end{equation}
}
\vspace{-4mm}

\noindent where $\widetilde{\nabla}_{m}$ is the stochastic gradient with respect to the parameters of modality $m$'s encoder. The classifier update is also modified as 
\begin{equation}\label{eq:update12}
   \hspace{-2mm} \bm{\varpi}^{(t), k+1}_{n} = \bm{\varpi}^{(t), k}_{n} - \eta^{(t),k} \hspace{.3mm}\Gamma_{\mathcal{M}_n,n}^{(t)} \widetilde{\nabla}_{\bm{\varpi}} \mathcal{L}_n(\bm{\omega}^{(t), k}_n; \mathcal{B}_n^{(t),k}),\hspace{-1.5mm}
\end{equation}
where $\widetilde{\nabla}_{\bm{\varpi}}$ is the stochastic gradient with respect to the classifier parameters.

To make DGB applicable for FL scenarios it should be adapted to one of the most unique aspects of FL settings, which is the existence of \textit{non-iid} and biased local datasets across the institutions.\footnote{This further results in local ML bias specifically when longer local round lengths $K$ are used.} In particular, obtaining \eqref{equ:o-def} and \eqref{equ:g-def} with naïve averaging of the local loss values of various modality combinations in the system renders the computed values ineffective under non-iid data. For example, consider an institution where the distribution of data can be biased (e.g., containing disproportionately high ratio of data from specific classes/labels, or containing disproportionately high ratio of data from specific patient demographics). This will make the local ML model of that specific institution overfit to the biased local data, which will result in a noticeably low local validation loss, severely impacting the aggregated losses used in DOGR upon naïve averaging of the loss values across institutions to obtain \eqref{equ:o-def} and \eqref{equ:g-def}. As a result, we have to migrate from conventional averaging of the losses across all institutions to the methods that can capture the quality of local datasets across institutions when aggregating their loss values.
We achieve this through introducing PCW, which controls the contribution of each institution's loss to the DGB parameter estimation based on the distance between its local data distribution $\mathcal{D}_n$ and the global data distribution $\mathcal{D}$. In doing so, in PCW, we will define a dynamic metric that conceptualizes the distance between institutions data distributions, and manipulates the impact with which each $\mathcal{L}_n^{\mathsf{Tr},(t)}$ and $\mathcal{L}_n^{\mathsf{Va},(t)}$ is taken into effect.
\vspace{-1mm}

\begin{remark}\label{remark}[Learning Methodology, Unbalanced Number of Modalities, Robustness to Non-iid data and Data Scarcity]
    The issue of varying convergence rates across modalities has also been observed in conventional centralized multi-modal ML \cite{gradient-blending}. To address this issue, \cite{gradient-blending} proposed a methodology, called gradient blending (GB), for centralized ML settings, which entails the use of a separate loss for each modality and a multi-modal model loss, weighting gradients non-uniformly when back-propagated over the ML model during training.
    As a result, since our methodology is inspired by \cite{gradient-blending}, although it contains multiple nuances and differences compared to the method developed in \cite{gradient-blending}, we call it \textit{Distributed Gradient Blending} (DGB). In particular, our method differs significantly from the naïve extension of GB to the FL setting as done in~\cite{mm-fl-gradblend}.
    In particular, existing methods \cite{mm-fl-gradblend, gradient-blending} use multiple sub-networks (i.e., one set of encoder and classifier for each modality), and thus, multiple losses to back-propagate during ML model training. In contrast, our work does not entail using multiple sub-networks; instead, we use only one classifier in each institution's ML model alongside separate encoders dedicated to various modalities. We further use only one loss function and weigh its resulting gradient vector elements based on the encoder they belong to. This reduces the computation complexity of our method and the number of model parameters to train in each institution. 
    Also, in addition to having a higher number of parameters, in the naïve implementations of gradient blending \cite{gradient-blending, mm-fl-gradblend}, the methods require the train data available ($\mathcal{D}_n^{\mathsf{Tr}}$) to be split into two sections, one of which is used for the calculation of the overfitting to generalization ratio. This raises challenges when facing a scenario with scarce data, which is frequent in medical domains given the labor needed to annotate data. Our method, on the other hand, uses system-wide characteristics and values of the losses to calculate the DOGR which removes the necessity to have each of the train and validation data to be split into two at each institution, enabling more compatibility with data-scarce scenarios. This is specifically shown in our method when using the averaged loss of the local ML models for each combination, instead of using the local loss for each local model separately.
    More importantly, we address the side-effects of non-iid data on the performance of multi-modal FL through PCW which is a new condition that arises because of the new configuration of DGB, aiming to elevate its performance.
\end{remark}

\vspace{-4mm}

\subsection{Proximity-Aware Client Weighting (PCW)}\label{sub:PCW}

We propose PCW to tackle the problem of non-iid distribution of local datasets, which leads to the bias of locally trained models in multi-modal FL. PCW defines a measure to weigh the institutions' local model losses (i.e.,  $\mathcal{L}_n^{\mathsf{Va},(t)}$ and $\mathcal{L}_n^{\mathsf{Tr},(t)}$ given in \eqref{equ:validation-loss} and \eqref{equ:train-loss}) to obtain the normalized losses (i.e., $\overline{\mathcal{L}}_C^{\mathsf{Va}, (t)}$ and $\overline{\mathcal{L}}_C^{\mathsf{Tr}, (t)}$ used in \eqref{equ:o-def} and \eqref{equ:g-def}, and subsequently \eqref{equ:original-OGR2}. This measure is defined based on the quality of local datasets of institutions. In order to obtain this measure, we rely on a key observation: considering that the ML models at institutions are initialized with a unique global model at the beginning of each local training round, the difference in their local gradient trajectories (i.e., their cumulative gradient generated during their local ML model training) is an indicator of the difference in their local data distributions.
Furthermore, the difference/closeness between the local gradient trajectories and the gradient trajectory of the global model is an indicator of the similarity of local datasets to the global dataset (i.e., institutions with higher similarity of local gradient trajectory to that of the global model have a closer local data distribution to that of global dataset, and thus a higher data quality). Using this measure, we will be able to calculate $\overline{\mathcal{L}}_C^{\mathsf{Va}, (t)}$ and $\overline{\mathcal{L}}_C^{\mathsf{Tr}, (t)}$, while explicitly considering the non-iid nature of data across the institutions. In particular,  $\overline{\mathcal{L}}_C^{\mathsf{Va}, (t)}$ and $\overline{\mathcal{L}}_C^{\mathsf{Tr}, (t)}$ are later computed through assigning a smaller weight to the loss of institutions with more biased (i.e., lower quality) datasets, while assigning a higher weight to the loss of institutions with higher quality data. More reliance/emphasis on the loss values of institutions with higher data quality through PCW, leads to a better capture of the behavior of each specific modality combination for the global distribution of data.

To realize PCW, we thus aim to compute the gradient trajectories at institutions and their similarities to that of global model. Therefore, we calculate the cumulative gradient of institution $n$'s model possessing modality combination $C=\mathcal{M}_n$ during the latest global aggregation step, denoted by $\Delta\bm{\omega}_{C, n}^{(t)}$, and then compute the trajectory of the global model gradient, denoted by $\Delta\bm{\omega}_{C,G}^{(t)}$. Mathematically, $\Delta\bm{\omega}_{C, n}^{(t)}$ and $\Delta\bm{\omega}_{C,G}^{(t)}$ are given by (note that $\bm{\omega}_{n}^{(t), 0} =  [\{\widehat{\bm{\omega}}_m^{(t)}\}_{m \in C}:\widehat{\bm{\varpi}}_{C}^{(t)}]$)
\begin{align} \label{equ:local-cumm-grad}
   \Delta \bm{\omega}_{C, n}^{(t)} & =  
     \Big(\bm{\omega}_{n}^{(t), 0} - \bm{\omega}_{n}^{(t), K} \Big)  = 
     \Big(\bm{\omega}_{n}^{(t), 0} - \bm{\omega}_{n}^{(t), 1} \Big) \nonumber\\
    &
    ~~~+ \Big(\bm{\omega}_{ n}^{(t), 1} - \bm{\omega}_{n}^{(t), 2} \Big) 
    + \cdots + \Big(\bm{\omega}_{n}^{(t), K-1} - \bm{\omega}_{n}^{(t), K} \Big)  
   \nonumber \\&= \sum_{j=0}^{K-1}\eta^{(t),j}\widetilde{\nabla}\mathcal{L}_n(\bm{\omega}^{(t), j}_n; \mathcal{B}_n^{(t),j}) 
\end{align}
and similarly
\begin{equation} \label{equ:global-grad}
   \hspace{-1.5mm} \Delta \bm{\omega}_{C,G}^{(t)}\hspace{-.5mm}= \hspace{-.5mm}
    \Big[\{\widehat{\bm{\omega}}_m^{(t)}\}_{m \in C}\hspace{-.5mm}:\hspace{-.5mm}\widehat{\bm{\varpi}}_{C}^{(t)}\Big]\hspace{-.5mm}-\hspace{-.5mm}\Big[\{\widehat{\bm{\omega}}^{(t+1)}_m\}_{m \in C}\hspace{-.5mm}:\hspace{-.5mm}\widehat{\bm{\varpi}}_{C}^{(t+1)}\Big].\hspace{-1mm}
\end{equation}
Next, to compute the similarity between the local and global gradient trajectories, we adopt the \textit{inner product} operator in defining a similarity/closeness metric as follows:
\begin{equation}\label{eq:innerProd}
    \rho_{C, n}^{(t)} = \langle \Delta \bm{\omega}_{C, n}^{(t)} , \Delta \bm{\omega}_{C,G}^{(t)}\rangle= \Delta \bm{\omega}_{C, n}^{(t)} \cdot \Delta \bm{\omega}_{C,G}^{(t)}.
\end{equation}
Considering~\eqref{eq:innerProd}, $\rho_{C, n}^{(t)}$ takes larger values when local gradient trajectory $\Delta \bm{\omega}_{C, n}^{(t)}$ exhibits a higher similarity to that of global gradient trajectory $\Delta\bm{\omega}_{C}^{(t)}$.\footnote{In addition to inner product, we also experimented with other metrics of similarity between the vectors, such as the infimum of the difference in the parameters. Nevertheless, inner product exhibited the best result in our experiments, and thus the rest of the metrics are omitted for brevity.}
To have comparable values of $\rho_{C, n}^{(t)}$ across institutions, we normalize the values obtained in~\eqref{eq:innerProd} to obtain a normalized positive similarity metric as follows:
\begin{equation} \label{equ:rho-bar}
    \overline{\rho}_{C,n}^{(t)} = {e^{\tau \rho_{C, n}^{(t)}}}\Big/ \Big({\sum_{n' \in \mathcal{N}} b_{C,n'}e^{\tau\rho_{C,n'}^{(t)}}}\Big),
\end{equation}
where $b_{C,n'}$ follows the same definition as in~\eqref{equ:aggregation-classifier} and $\tau$ is a tunable temperature (chosen as $\tau = 1$ in our simulations).

Having the train and validation losses, i.e.,~\eqref{equ:validation-loss} and~\eqref{equ:train-loss},
for each institution possessing subset $C$ of modalities, we then derive the adjusted average train and validation loss of the respective subset of modalities in the system at the end of each global aggregation step $t$, denoted by $\overline{\mathcal{L}}^{\mathsf{Tr}, (t)}_C$ and $\overline{\mathcal{L}}^{\mathsf{Va}, (t)}_C$, respectively, via the integration of the above similarity metric as follows ($b_{C,n}$ is defined the same as in \eqref{equ:aggregation-classifier}):
\begin{equation}
\begin{aligned} \label{equ:new-ogr-bars}
    \overline{\mathcal{L}}^{\mathsf{Tr},(t)}_C &= \frac{1}{\sum_{n \in \mathcal{N}} b_{C,n}} \sum_{n \in \mathcal{N}} \overline{\rho}_{C,n}^{(t)} b_{C,n} \mathcal{L}^{\mathsf{Tr},(t)}_n, \\ 
    \overline{\mathcal{L}}^{\mathsf{Va},(t)}_C &= \frac{1}{\sum_{n \in \mathcal{N}} b_{C,n}} \sum_{n \in \mathcal{N}} \overline{\rho}_{C,n}^{(t)} b_{C,n} \mathcal{L}^{\mathsf{Va},(t)}_n.
\end{aligned}
\end{equation}

Using the estimates in \eqref{equ:new-ogr-bars}, we can substitute the values for $\overline{\mathcal{L}}^{\mathsf{Tr},(t)}_C$ and $\overline{\mathcal{L}}^{\mathsf{Va},(t)}_C$ in \eqref{equ:o-def} and \eqref{equ:g-def} to calculate the overfitting and generalization, and consequently we can calculate the weights in \eqref{equ:original-OGR2}. Compared to a naïve averaging of the training and validation losses, this new formulation of $\overline{\mathcal{L}}^{\mathsf{Tr},(t)}_C$ and $\overline{\mathcal{L}}^{\mathsf{Va},(t)}_C$ enhances the performance by weighing losses from institutions with closer data distribution to the global data distribution $\mathcal{D}$ with higher values, thus tackling the challenge of non-iid distributed data when calculating the DOGR coefficients. 
This is an important step since the average losses being calculated in \eqref{equ:o-def} and \eqref{equ:g-def} will be used for the entire system in~\eqref{equ:original-OGR2}. Therefore, the institutions where data induces more bias should be given a smaller weight/importance when calculating the average loss, which will determine the DGB weights for all institutions. Our overall methodology for multi-modal FL comprising DGB and PCW is provided in Algorithm~\ref{alg:main}.

\vspace{-3mm}
\section{Numerical Experiments}
\label{sec:experiments}

\noindent 
In this section, we first detail the datasets in Sec.~\ref{subsec:data}.  Sec.~\ref{subsec:preprocessing} outlines the preprocessing steps undertaken for each data modality. We then review the baseline methods, describing their workings, result replication, and their relevance to our scenario in Sec.~\ref{subsec:baselines}. Sec.~\ref{subsec:hyperparameters} describes the tuning of hyperparameters. Finally, simulation results are presented in Sec.~\ref{subsec:results}, complemented with discussions  on their implications.

\begin{algorithm}[t]
    \caption{Multi-modal FL over Institutions with Unbalanced Modalities}\label{alg:main}
 	\SetKwFunction{Union}{Union}\SetKwFunction{FindCompress}{FindCompress}
 	\SetKwInOut{Input}{input}\SetKwInOut{Output}{output}
 	 	{\footnotesize
 	\Input{Institutions $\mathcal{N}$, Datasets of institutions $\{\mathcal{D}_{n}\}_{n\in\mathcal{N}}$, Number of global steps $T$, Number of local SGD iterations $K$ in each global step, Initial DOGR weights for modality encoders $\Gamma^{(0)}_{m,n}$, Initial DOGR weights for institution classifiers  $\Gamma^{(0)}_{\mathcal{M}_n,n}$}
 	\For{global aggregation step $t=1$ to $T$}{
     {\color{blue}\% On the institutions' Side} \\
 	 \For{institution $n=1$ to $|\mathcal{N}|$ in parallel}{
        \If{$t < 2$}{
            Set encoder of modality $m$'s learning rate coefficients  
            $\Gamma^{(t)}_{m,n} = \Gamma^{(0)}_{m,n}$\\
            Set the classifier's learning rate coefficients  $\Gamma^{(t)}_{\mathcal{M}_n,n} = \Gamma^{(0)}_{\mathcal{M}_n,n}$
        }
        \Else{Calculate the encoder and classifier learning rate coefficients $\Gamma_{m,n}^{(t)}, \forall m \in \mathcal{M}_n$ and $\Gamma_{\mathcal{M}_n,n}^{(t)}$ using \eqref{equ:original-OGR2}}
        Perform $K$ rounds of local SGD based on  \eqref{eq:update11} and \eqref{eq:update12}. \\
        Send the train loss $\mathcal{L}_n^{\mathsf{Tr}}(\bm{\omega}_n^{(t), K})$, validation loss $\mathcal{L}_n^{\mathsf{Va}}(\bm{\omega}_n^{(t), K})$, and the model parameters  $\bm{\omega}_n^{(t), K}=\big\{\{\bm{\omega}_{m,n}^{(t),K}\}_{m \in \mathcal{M}_n}: \bm{\varpi}^{(t),K}_n\big\}$ to the server 
     }
      {\color{red}\% On the Server's Side}\\
     \For{modality $m=1$ to $|\mathcal{M}|$}{
        Aggregate the encoders with parameters $\bm{\omega}_{m,n}^{(t),K}$ based on  \eqref{equ:aggregation-encoder} to obtain $\widehat{\bm{\omega}}_{m}^{(t+1)}$.
     }
     \For{{combination $C \in \mathcal{P}(\mathcal{M})$}}{
        Aggregate the classifiers with parameters $\bm{\varpi}_{n}^{(t), K}$ based on  \eqref{equ:aggregation-classifier} to obtain $\widehat{\bm{\varpi}}_{C}^{(t+1)}$.
     }
     \For{modality combination $C \in \mathcal{P}(\mathcal{M})$}{
     Calculate the weight $\overline{\rho}_{C,n}^{(t)}$ based on \eqref{equ:rho-bar}
     
        \If{$t \geq 2$}{
            Calculate the average training and validation losses in global aggregation step $t$, $\overline{\mathcal{L}}^{\mathsf{Va}, (t)}_C$ and $\overline{\mathcal{L}}^{\mathsf{Tr}, (t)}_C$ based on \eqref{equ:new-ogr-bars}\\
            Calculate $O_C^{(t)}$ and $G_C^{(t)}$ based on \eqref{equ:o-def} and \eqref{equ:g-def}
        }
     }
     Send the calculated $\widehat{\bm{\omega}}_{m}^{(t+1)}$, $\widehat{\bm{\varpi}}_{C}^{(t+1)}$, $O_C^{(t)}$ and $G_C^{(t)}, C\in\mathcal{P}(\mathcal{M})$ back to the institutions.
 	  }
 	  }
 	  \vspace{-.1mm}
  \end{algorithm}

\vspace{-3mm}
\subsection{Dataset and Data Distribution}\label{subsec:data}
We use datasets from The Cancer Genome Atlas program (TCGA) \cite{Zhu2014} for three cancer types, namely, Breast Invasive Carcinoma (BRCA), Lung Squamous Cell Carcinoma (LUSC), and Liver Hepatocellular Carcinoma (LIHC). From each cohort, we will use three of the available modalities: \textit{mRNA sequence}, \textit{histopathological images}, and \textit{clinical information}. 
The data from these cohorts is filtered to keep only those patient data points that have all three modalities to make the results comparable to centralized ML as conducted in \cite{SHAO2020101795} and the FL settings described in \cite{mm-fl-gradblend,XIONG2022110}.
We use this data to simulate each institution's local dataset in our experiments. 
The distribution of the class labels across the cohorts is summarized in Table \ref{tab:data-dist}.

\begin{table*}[!h]
\vspace{1.5mm}
\caption{Available patients for each stage across the cohorts used for binary and multi-class classification.}
\vspace{-1.5mm}
\centering
\label{tab:data-dist}
{\notsotiny
\begin{tabularx}{\textwidth}{ 
  | > {\centering\arraybackslash}m{19.21mm} 
  || >{\centering\arraybackslash}m{17mm} 
  | >{\centering\arraybackslash}m{17mm} 
  | >{\centering\arraybackslash}m{17mm} 
  | >{\centering\arraybackslash}X
  || >{\centering\arraybackslash}X|} 
  \hline
          \rowcolor{LightBlue}
          \diagbox[width=23mm]{Stage}{Cohort\hspace{-2mm}} & BRCA & LUSC & LIHC & \makecell[cc]{All Cohorts \\ (Binary/\textbf{Multi-Class})} & \makecell[cc]{Stage Ratio (\%) \\ (Binary/\textbf{Multi-Class})}\\
          \hline\hline
          Stage I & 155 & 152 & 152 & 459 & $\simeq$40.55/\textbf{31.48} \\
          \hline
          \rowcolor{LightBlue}
          Stage II & 488 & 108 & 77 & 673 & $\simeq$59.45/\textbf{46.16} \\
          \hline
          Stage III & 208 & 44 & 74 & 326 & $\simeq$N.A./\textbf{22.36} \\
          \hline
          \rowcolor{LightBlue}
          All Stages  (Binary/\textbf{Multi-Class}) &
          643/\textbf{851} &
          260/\textbf{304} &
          229/\textbf{303} &
          1132/\textbf{1458} \\
          \hhline{*5{-}|~}
          \noalign{\vspace{1pt}}
          \hhline{*4{-}|~}
          Cohort Ratio (\%) (Binary/\textbf{Multi-Class}) &
          $\simeq$56.80/\textbf{58.37} & $\simeq$22.97/\textbf{20.85} & $\simeq$20.23/\textbf{20.78} \\
          \hhline{*4{-}|~}
\end{tabularx}
}
\vspace{-4mm}
\end{table*}

We next distribute the refined dataset across $21$ institutions, where, to emulate real-world settings, we consider the existence of unbalanced modality combinations across the institutions. In particular, the configuration of modalities in each institution's local dataset is summarized in Table~\ref{tab:combination-info}. Given both 
binary classification \cite{sarkar2023,litjens2016deep,hadjiyski2020kidney,SHAO2020101795} and multi-class classification \cite{detect-lc-22,Xiehist19} are considered in the existing literature on the task of cancer staging,
to provide benchmarks for the follow-up works in the emerging area of multi-modal FL, we consider both binary and multi-class classification and present their respective results in this section.

Next, we further delve into the reflection of non-iid-ness/heterogeneity of data across the institutions. In particular, in our network of interest with multiple data modalities and various stages of cancer (i.e., data labels),  non-iid-ness/heterogeneity of data can be interpreted at two levels: (i) type-based heterogeneity, and (ii) class-based heterogeneity, as elaborated below.

\begin{table}[b]
\vspace{2mm}
\caption{Modality combinations used across institutions.}
\centering
\label{tab:combination-info}
{\scriptsize
\begin{tabularx}{0.49\textwidth}{ 
  | >{\centering\arraybackslash}X
  | >{\centering\arraybackslash}X
  | >{\centering\arraybackslash}X
  || >{\centering\arraybackslash}X|}
    \hline
   \rowcolor{DarkBlue} \multicolumn{3}{|c||}{Data Modalities} & Entity \\
    \hhline{|*3{-}||*1{-}|}
    \rowcolor{LightBlue} mRNA & Image & Clinical & Institution \# \\
    \hline
    \hline
    %% Type 1 institutions
    \checkmark & \checkmark & \checkmark & {1, 2, 3}\\
    \hline
    \checkmark & \checkmark & & {4, 5, 6}\\
    \hline
    \checkmark & & \checkmark & {7, 8, 9}\\
    \hline
    & \checkmark & \checkmark & {10, 11, 12}\\
    \hline
    \checkmark & & & {13, 14, 15}\\
    \hline
    & \checkmark & & {16, 17, 18}\\
    \hline
    & & \checkmark & {19, 20, 21}\\
    \hline
\end{tabularx}
}
\end{table}

\begin{itemize}[leftmargin=4mm]
    \item \textbf{Type-based Heterogeneity:} We keep the same ratio of class labels across institutions' local datasets; $\mathcal{D}_n~\forall n$ contain the same fraction of data points of different labels. However, the ratio of the data points belonging to various cohorts will be different across institutions, which leads to a bias toward a specific cancer \textit{type} in each institution's local data. 
    Following the modality spread across the institutions given in Table \ref{tab:combination-info}, the fraction of each cohort's data inherited by each institution is given in  Table \ref{tab:type-heterogeneity-distr} for each of the binary and multi-class classification scenarios.
    \item \textbf{Class-based Heterogeneity:} Institutions' local datasets will have the same ratio of datapoints of the three cancer types. However, institutions' local datasets contain a varying ratio of datapoints belonging to different \textit{classes (i.e., cancer stages)}, which leads to a bias toward a specific cancer \textit{stage/class} in each institution's local data. The class distribution of data for binary and multi-class classification cases is shown in Table \ref{tab:class-heteogeneity-distr}.
\end{itemize}

\begin{table*}[t]
\caption{Type-based heterogeneity data distribution across institutions based on the fraction of institutions' local dataset consisting of each cohort's data.\vspace{-1mm}}
\centering
\label{tab:type-heterogeneity-distr}
{\notsotiny
\begin{tabularx}{0.979\textwidth}{ 
      | >{\centering\arraybackslash}X
      | >{\centering\arraybackslash}X
      | >{\centering\arraybackslash}X
      | >{\centering\arraybackslash}m{1mm}
      | >{\centering\arraybackslash}X
      | >{\centering\arraybackslash}X
      | >{\centering\arraybackslash}X
      || >{\centering\arraybackslash}m{19mm}
      | >{\centering\arraybackslash}m{14mm}|}
        \hline
        \rowcolor{DarkBlue} \multicolumn{3}{|c|}{ Class Labels (Binary Classification)}  &\cellcolor{white} & \multicolumn{3}{c||}{Class Labels (Multi-Class Classification)} & \multicolumn{2}{c|}{Entity}\\
        \hhline{|*3{-}|*1{~}|*3{-}||*2{-}|}
        \rowcolor{LightBlue}
        BRCA & LUSC & LIHC & \cellcolor{white} & BRCA & LUSC & LIHC & {Institution \#} & {Category}\\
        \hline
        \hline
        %% Type 1 institutions
        $\simeq$ 56\% & $\simeq$ 22\% & $\simeq$ 20\% &
        \cellcolor{white} &
        $\simeq$ 58\% & $\simeq$ 21\% & $\simeq$ 21\% &
        { 1, 4, 7, 10, 13, 16, 19} & 1\\
        \hhline{|*3{-}|*1{~}|*3{-}||*2{-}|}
        %% Type 2 institutions
        \rowcolor{LightBlue}
        $\simeq$ 54\% & $\simeq$ 10\% & $\simeq$ 35\% &
        \cellcolor{white} &
        $\simeq$ 48\% & $\simeq$ 28\% & $\simeq$ 24\% &
        { 2, 5, 8, 11, 14, 17, 20} & 2\\
        \hhline{|*3{-}|*1{~}|*3{-}||*2{-}|}
        %% Type 3 institutions
        $\simeq$ 58\% & $\simeq$ 35\% & $\simeq$ 6\% &
        \cellcolor{white} &
        $\simeq$ 62\% & $\simeq$ 12\% & $\simeq$ 26\% &
        {3, 6, 9, 12, 15, 18, 21} & 3\\
        \hline
    \end{tabularx}
}
\end{table*}

\begin{table*}[t]
\vspace{-1mm}
\caption{Class-based heterogeneity data distribution across institutions based on the fraction of institutions' local dataset consisting of each class label.}
\vspace{-2mm}
\centering
\label{tab:class-heteogeneity-distr}
{\notsotiny
\begin{tabularx}{0.99\textwidth}{ 
  | >{\centering\arraybackslash}X
  | >{\centering\arraybackslash}X
  | >{\centering\arraybackslash}m{0.1mm}
  | >{\centering\arraybackslash}X
  | >{\centering\arraybackslash}X
  | >{\centering\arraybackslash}X
  || >{\centering\arraybackslash}X
  | >{\centering\arraybackslash}m{14mm}|}
    \hline
    \rowcolor{DarkBlue} \multicolumn{2}{c|}{Class Labels (Binary Classification)}&\cellcolor{white} &\multicolumn{3}{|c||}{ Class Labels (Multi-Class Classification)} & \multicolumn{2}{c|}{Entity}\\
    \hhline{|*2{-}|~|*3{-}||*2{-}|}
    \rowcolor{LightBlue} Stage I& Stage II & \cellcolor{white} & Stage I & Stage II & Stage III & {Institution \#} & Category\\
    \hline
    \hline
    %% Type 1 institutions
    $\simeq$ 41\% & $\simeq$ 59\% &
    \cellcolor{white} &
    $\simeq$ 32\% & $\simeq$ 46\% & $\simeq$ 22\% &
    { 1, 4, 7, 10, 13, 16, 19} & 1\\
    \hhline{|*2{-}|~|*3{-}||*2{-}|}
    %% Type 2 institutions
    \rowcolor{LightBlue}
    $\simeq$ 62\% & $\simeq$ 38\% &
    \cellcolor{white} &
    $\simeq$ 36\% & $\simeq$ 34\% & $\simeq$ 30\% &
    { 2, 5, 8, 11, 14, 17, 20} & 2\\
    \hhline{|*2{-}|~|*3{-}||*2{-}|}
    %% Type 3 institutions
    $\simeq$ 19\% & $\simeq$ 81\% &
    \cellcolor{white} &
    $\simeq$ 30\% & $\simeq$ 54\% & $\simeq$ 16\% &
    {3, 6, 9, 12, 15, 18, 21} & 3\\
    \hline
\end{tabularx}
}
\vspace{-5mm}
\end{table*}

\textbf{Heterogeneity Severity Comparison.} It is worth mentioning that type-based heterogeneity can be though of as a \textit{more fundamental} level of heterogeneity in which the institutions local datasets have unbalanced ratios of the cancer types as compared to the unbalanced ratio of their labels imposed in class-based heterogeneity. As a result, type-based heterogeneity will have a more pronounced impact on the model training since during the local model training at each institution, for each cancer type, a specific part of the input information\footnote{For example, specific elements in the mRNA sequence, specific extracted image features, specific clinical information} is important: having unbalanced ratios of cancer types may lead to the confusion of the local model regarding the identification of the most important parts of the input information to focus for conducting classification. 
Nevertheless, since the decisive input features for classification may be shared across different classes for the same cancer type, imposing class-based heterogeneity while having the same distribution of cancer types across institutions will not result in such a severe bias in the local ML models during training.
For example, to differentiate between different stages of a cancer the model may focus on the same feature set of the histopathological images, which may correspond to the location of a sensitive tissue. Nevertheless, such feature set (or the location of focus) may differ across different cancers. Thus, having unbalanced ratios of cancer types may bias the model towards focusing on the input features of the majority type of the local datapoints, which may differ across cancer types; however, having unbalanced classes may not impact the local model bias as much.

%when differentiating between different classes (i.e., stages), often the most important features in classification are recognized -- simply put, higher weights in the input are used to feed the important feature values to the deeper layers of the ML model -- and the classification is improved through determining the relationship between those features by iteratively refining the ML model weights. 

All experiments were conducted on a server with NVIDIA's Tesla A-100 GPUs. For software implementation, we used PyTorch \cite{NEURIPS2019_9015} module using Python programming language. Also, our codes are available at \href{https://github.com/KasraBorazjani/dgb-pcw-fl}{https://github.com/KasraBorazjani/dgb-pcw-fl.git}.

\vspace{-3.5mm}
\subsection{Preprocessing Phase}\label{subsec:preprocessing}

\begin{figure}
    \centering
    \includegraphics[width=0.44\textwidth]{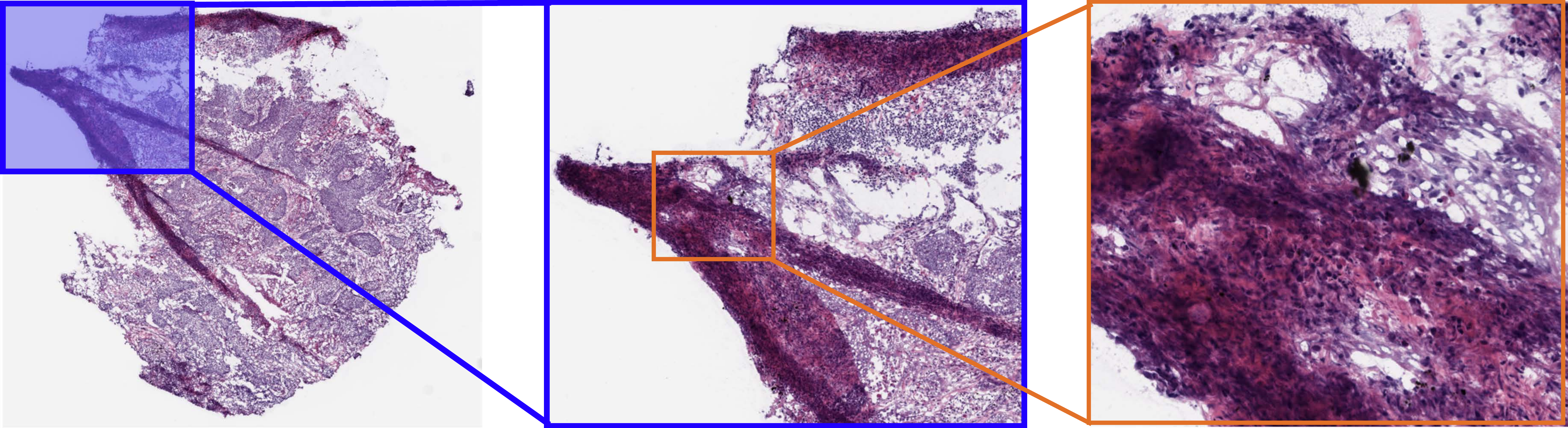}
    \caption{A sample of the histopathological image, containing dense pixels.}
    \label{fig:wsi-zoom}
\end{figure}

As dealing with data in raw format is highly challenging, especially in the case of histopathological images (see Fig. \ref{fig:wsi-zoom}) where data is of large size, we follow the method of \cite{Jun-hist-fe} to preprocess data. In this procedure, the nucleus segmentation is done using the unsupervised method of \cite{phoulady-hist-fe}. We then extract $10$ cell-level features, namely, (i) nucleus area, (ii) major axis length, (iii) minor axis length, (iv) major to minor axis ratio, (v) mean distance to the nearby nuclei, (vi) maximum distance to the nearby nuclei, (vii) minimum distance to the nearby nuclei, and the average value of each of the (viii) red, (ix) green, and (x) blue channels in the surrounding area of the nuclei. These will constitute our \textit{cell-level} features. To make a \textit{patient-level} or \textit{slide-level} representation for each patient, we take each of  cell-level features and create a $10$-bin histogram for it, alongside the five metrics of mean, standard deviation, skewness, kurtosis, and entropy. Taking each bin of the histogram and the aforementioned metrics  as a \textit{patient-level} feature will yield $150$ features in total which we use for describing a single histopathological image (WSI) for each patient. The rest of data modalities (i.e., \textit{mRNA sequence} and \textit{clinical information}) are fed to the ML model in raw formats.

\vspace{-3.2mm}
\subsection{Baseline Methods}\label{subsec:baselines}
\vspace{-.2mm}
We compare the results of our method with three baselines:
\begin{enumerate}[leftmargin=4.5mm]
    \item \textit{Conventional Multi-Modal FL (CM-FL):} Following the model in Fig.~\ref{fig:system-model}, institutions have heterogeneous modalities and local ML models are trained using the same learning-rate and aggregated using the FedAvg algorithm \cite{XIONG2022110}.
    \item \textit{Multi-Modal FL with Hierarchical Gradient Blending (HGB-FL):} We consider \cite{mm-fl-gradblend} to test our method versus another implementation of gradient bending for FL. This approach splits the train data at each institution to calculate the gradient blending weights and tunes the weights with which the ML models will be aggregated at the server.    
    \item \textit{CM-FL + DGB:} The institutions will possess heterogeneous modalities and their ML models will be trained using DGB without the use of PCW.
    \item \textit{UM-FL}: The upper bound FL performance is achieved when all institutions possess all of data modalities (i.e., the scenario resembles uniform-modality distribution), removing the barrier of different convergence rates across  modalities. This baseline is only added to provide an estimate of the achievable performance.
\end{enumerate}

\begin{table}[!b]
\vspace{1mm}
\caption{Different architectures for the mRNA encoder.}
\vspace{-2mm}
\centering
\label{tab:mrna-model-depth}
{\scriptsize
\begin{tabularx}{0.49\textwidth}{ 
  | >{\centering\arraybackslash}X 
  || >{\centering\arraybackslash}X
  | >{\centering\arraybackslash}X
  | >{\centering\arraybackslash}X |} 
  \hline
  \rowcolor{LightBlue}   Layer Type & \multicolumn{3}{c|}{Model Depth} \\
  \hhline{|*1{-}||*3{-}|}
    & Small & Medium & Large \\
  \hline\hline
   \rowcolor{LightBlue}   Input Layer & \multicolumn{3}{c|}{FC (20531, ReLU)} \\
  \hline
  Layer 1 & FC (4096, ReLU) & FC (8192, ReLU) & FC~(16384,~ReLU) \\
  \hline
   \rowcolor{LightBlue}  Layer 2 & FC (2048, ReLU) & FC (4096, ReLU) & FC (8192, ReLU) \\
  \hline
  Layer 3 & FC (512, ReLU) & FC (2048, ReLU) & FC (4096, ReLU) \\
  \hline
   \rowcolor{LightBlue}  Layer 4 & FC (128, ReLU) & FC (512, ReLU) & FC (2048, ReLU) \\
  \hline
  Layer 5 & FC (64, ReLU) & FC (128, ReLU) & FC (512, ReLU) \\
  \hline
   \rowcolor{LightBlue}  Layer 6 & FC (32, ReLU) & FC (64, ReLU) & FC (128, ReLU) \\
  \hline
  Layer 7 & --- & FC (32, ReLU) & FC (64, ReLU) \\
  \hline
   \rowcolor{LightBlue}  Layer 8 & --- & --- & FC (32, ReLU) \\
  \hline 
 {Output~Layer  \hspace*{-\baselineskip}{\notsotiny  Binary/\textbf{Multi-Class}}\hspace*{-\baselineskip}} & \multicolumn{3}{c|}{FC (2, LogSoftmax) / \textbf{FC (3, LogSoftmax)}}\\
  \hline
\end{tabularx}
}
\end{table}

\begin{table*}[ht]
\caption{Accuracy (\%) of different methods throughout the global aggregations under both binary and multi-class classification for type-based heterogeneity.}
\vspace{-1.5mm}
\centering
\label{tab:accuracy-table-type-based}
{\scriptsize
\begin{tabularx}{\textwidth}{ 
  | >{\centering\arraybackslash}m{1.2cm}
  || >{\centering\arraybackslash}m{0.9cm}
  | >{\centering\arraybackslash}m{0.9cm}
  | >{\centering\arraybackslash}m{1.1cm}
  | >{\centering\arraybackslash}m{1.1cm}
  || >{\centering\arraybackslash}m{0.9cm}
  | >{\centering\arraybackslash}m{0.1mm}
  | >{\centering\arraybackslash}m{1.2cm}
  || >{\centering\arraybackslash}m{0.9cm}
  | >{\centering\arraybackslash}m{0.9cm}
  | >{\centering\arraybackslash}X
  | >{\centering\arraybackslash}m{1.1cm}
  || >{\centering\arraybackslash}m{0.9cm}|} 
  
    \hline
     \rowcolor{DarkBlue}\multicolumn{6}{|c|}{Binary Classification} &  { \cellcolor{white}}   & \multicolumn{6}{c|}{Multi-Class Classification}\\
    % \hhline{|~~~~~~|*1{-}|~~~~~~|}
    \hhline{|*6{-}|*1{~}|*6{-}|}
    % \hhline{|*6{-}|*1{>{\arrayrulecolor[gray]{.8}}-}|*6{>{\arrayrulecolor[black]{.8}}-}|}
    % \hhline{|~~~~~~|*1{-}|~~~~~~|}
   \rowcolor{LightBlue}  \vspace{-2mm}\diagbox[height=8mm, width=16.25mm]{\tiny \hspace{-1.25mm}Global Round}{ \tiny Method}
    & {\vspace{-3mm} CM-FL} & HGB-FL & CM-FL + DGB & Our Method & UM-FL (Upperbound) &{ \cellcolor{white}}\multirow{-2}{*}{} & \vspace{-2mm}\diagbox[height=8mm, width=16.25mm]{\tiny \hspace{-1.25mm}Global Round}{ \tiny Method} & {\vspace{-3mm} CM-FL} & HGB-FL & CM-FL + DGB & Our Method & UM-FL (Upperbound) \\
    \hline
    \hline
    0 & $\simeq$ 52.34 & $\simeq$ 52.34 & $\simeq$ 52.34 & $\simeq$ 52.34 & $\simeq$ 52.34 &{ \cellcolor{white}} &
    0 & $\simeq$ 36.52 & $\simeq$ 36.52 & $\simeq$ 36.52 & $\simeq$ 36.52 & $\simeq$ 36.52\\
    
    \hhline{|*6{-}|*1{~}|*6{-}|}
    
    \rowcolor{LightBlue}
    100 & $\simeq$ 62.92 & $\simeq$ 62.39 & $\simeq$ \textbf{63.34} & $\simeq$ 61.80 & $\simeq$ 69.32 &
    { \cellcolor{white}} &
    100 & $\simeq$ 41.69 & $\simeq$ 49.32 & $\simeq$ 46.26 & $\simeq$ \textbf{49.93} & $\simeq$ 54.74\\
    
    \hhline{|*6{-}|*1{~}|*6{-}|}
    
    200 & $\simeq$ 59.55 & $\simeq$ 63.25 & $\simeq$ 63.64 & $\simeq$ \textbf{64.04} & $\simeq$ 73.38 &
    { \cellcolor{white}} &
    200 & $\simeq$ 44.26 & $\simeq$ 55.14 & $\simeq$ 48.25 & $\simeq$ \textbf{58.63} & $\simeq$ 63.18\\
    
    \hhline{|*6{-}|*1{~}|*6{-}|}
    
    \rowcolor{LightBlue}
    300 & $\simeq$ 62.92 & $\simeq$ 64.10 & $\simeq$ 64.24 & $\simeq$ \textbf{68.54} & $\simeq$ 76.24 &
    { \cellcolor{white}} &
    300 & $\simeq$ 46.58 & $\simeq$ 56.17 & $\simeq$ 49.63 & $\simeq$ \textbf{64.32} & $\simeq$ 68.25 \\
    
    \hhline{|*6{-}|*1{~}|*6{-}|}
    
    400 & $\simeq$~61.80 & $\simeq$ 65.81 & $\simeq$ 66.43 & $\simeq$ \textbf{71.91} & $\simeq$ 79.46 &
    { \cellcolor{white}} &
    400 & $\simeq$ 47.52 & $\simeq$ 56.83 & $\simeq$ 52.36 & $\simeq$ \textbf{66.24} & $\simeq$ 72.85 \\
    
    \hhline{|*6{-}|*1{~}|*6{-}|}
    
    \rowcolor{LightBlue}
    500 & $\simeq$ 64.04 & $\simeq$ 67.52 & $\simeq$ 67.13 & $\simeq$ \textbf{74.16} & $\simeq$ 79.46 &
    { \cellcolor{white}} &
    500 & $\simeq$ 47.82 & $\simeq$ 57.14 & $\simeq$ 53.64 & $\simeq$ \textbf{67.35} & $\simeq$ 74.47 \\
    
    \hhline{|*6{-}|*1{~}|*6{-}|}
    
    600 & $\simeq$ 61.80 & $\simeq$ 68.64 & $\simeq$ 69.21 & $\simeq$ \textbf{74.16}  & $\simeq$ 80.24 &
    { \cellcolor{white}} &
    600 & $\simeq$ 49.49 & $\simeq$ 57.52 & $\simeq$ 53.83 & $\simeq$ \textbf{67.84} & $\simeq$ 75.36 \\
    
    \hline
  
\end{tabularx}
}
\end{table*}

\begin{table*}[ht]
\caption{Accuracy (\%) of different methods throughout the global aggregations under both binary and multi-class classification for class-based heterogeneity.}
\vspace{-1.5mm}
\centering
\label{tab:accuracy-table-class-based}
{\scriptsize
\begin{tabularx}{\textwidth}{ 
  | >{\centering\arraybackslash}m{1.2cm}
  || >{\centering\arraybackslash}m{0.9cm}
  | >{\centering\arraybackslash}m{0.9cm}
  | >{\centering\arraybackslash}m{1.1cm}
  | >{\centering\arraybackslash}m{1.1cm}
  || >{\centering\arraybackslash}m{0.9cm}
  | >{\centering\arraybackslash}m{0.1mm}
  | >{\centering\arraybackslash}m{1.2cm}
  || >{\centering\arraybackslash}m{0.9cm}
  | >{\centering\arraybackslash}m{0.9cm}
  | >{\centering\arraybackslash}X
  | >{\centering\arraybackslash}m{1.1cm}
  || >{\centering\arraybackslash}m{0.9cm}|} 
  
    \hline
     \rowcolor{DarkBlue}\multicolumn{6}{|c|}{Binary Classification} &  { \cellcolor{white}}   & \multicolumn{6}{c|}{Multi-Class Classification}\\
    % \hhline{|~~~~~~|*1{-}|~~~~~~|}
    \hhline{|*6{-}|*1{~}|*6{-}|}
    % \hhline{|*6{-}|*1{>{\arrayrulecolor[gray]{.8}}-}|*6{>{\arrayrulecolor[black]{.8}}-}|}
    % \hhline{|~~~~~~|*1{-}|~~~~~~|}
   \rowcolor{LightBlue}  \vspace{-2mm}\diagbox[height=8mm, width=16.25mm]{\tiny \hspace{-1.25mm}Global Round}{ \tiny Method}
    & {\vspace{-3mm} CM-FL} & HGB-FL & CM-FL + DGB & Our Method & UM-FL (Upperbound) &{ \cellcolor{white}}\multirow{-2}{*}{} & \vspace{-2mm}\diagbox[height=8mm, width=16.25mm]{\tiny \hspace{-1.25mm}Global Round}{ \tiny Method} & {\vspace{-3mm} CM-FL} & HGB-FL & CM-FL + DGB & Our Method & UM-FL (Upperbound) \\
    \hline
    \hline
    0 & $\simeq$ 52.34 & $\simeq$ 52.34 & $\simeq$ 52.34 & $\simeq$ 52.34 & $\simeq$ 52.34 &
    { \cellcolor{white}} &
    0 & $\simeq$ 36.52 & $\simeq$ 36.52 & $\simeq$ 36.52 & $\simeq$ 36.52 & $\simeq$ 36.52\\
    
    \hhline{|*6{-}|*1{~}|*6{-}|}
    
    \rowcolor{LightBlue}
    100 & $\simeq$ 53.58 & $\simeq$ 63.14 & $\simeq$ \textbf{65.46} & $\simeq$ 63.48 & $\simeq$ 70.58 &
    {\cellcolor{white}} &
    100 & $\simeq$ 43.14 & $\simeq$ 51.37 & $\simeq$ 46.83 & $\simeq$ \textbf{55.32} & $\simeq$ 56.82\\
    \hhline{|*6{-}|*1{~}|*6{-}|}
   
    200 & $\simeq$ 57.13 & $\simeq$ 64.35 & $\simeq$ \textbf{67.18} & $\simeq$ {66.80} & $\simeq$ 75.23 &
    { \cellcolor{white}} &
    200 & $\simeq$ 47.23 & $\simeq$ 56.29 & $\simeq$ 53.82 & $\simeq$ \textbf{60.82} & $\simeq$ 63.43\\
    
    \hhline{|*6{-}|*1{~}|*6{-}|}
    
    \rowcolor{LightBlue} 
    300 & $\simeq$ 62.43 & $\simeq$ 64.66 & $\simeq$ 68.12 & $\simeq$ \textbf{68.23} & $\simeq$ 78.45 &
    { \cellcolor{white}} &
     300 & $\simeq$ 47.23 & $\simeq$ 58.34 & $\simeq$ 55.72 & $\simeq$ \textbf{64.17} & $\simeq$ 69.72 \\
    
    \hhline{|*6{-}|*1{~}|*6{-}|}
    
    400 & $\simeq$~65.24 & $\simeq$ 66.73 & $\simeq$ 69.57 & $\simeq$ \textbf{73.28} & $\simeq$ 79.75 &
    { \cellcolor{white}} &
    400 & $\simeq$ 47.82 & $\simeq$ 62.04 & $\simeq$ 51.93 & $\simeq$ \textbf{66.82} & $\simeq$ 72.55 \\
    
    \hhline{|*6{-}|*1{~}|*6{-}|}
    
    \rowcolor{LightBlue}
    500 & $\simeq$ 64.04 & $\simeq$ 68.65 & $\simeq$ 70.43 & $\simeq$ \textbf{76.24} & $\simeq$ 80.34&
    { \cellcolor{white}} &
    500 & $\simeq$ 48.53 & $\simeq$ 63.18 & $\simeq$ 56.72 & $\simeq$ \textbf{68.39} & $\simeq$ 74.28 \\
    
    \hhline{|*6{-}|*1{~}|*6{-}|}
    
    600 & $\simeq$ 62.84 & $\simeq$ 69.15 & $\simeq$ 71.13 & $\simeq$ \textbf{76.31}  & $\simeq$ 81.26&
    { \cellcolor{white}} &
    600 & $\simeq$ 48.53 & $\simeq$ 63.56 & $\simeq$ 57.26 & $\simeq$ \textbf{69.92} & $\simeq$ 75.10 \\
    \hline
  
\end{tabularx}
}
\end{table*}

\begin{table*}[ht]
\scriptsize
\caption{Global model accuracy (\%) for each dominant modality scenario for both binary classification and multi-class classification cases in the presence of type-based heterogeneity in local institution datasets.}
\vspace{-1.5mm}
\centering
\label{tab:non-uniform-level-type-based}
{
\begin{tabularx}{\textwidth}{ 
  | >{\centering\arraybackslash}m{19.85mm} 
  || >{\centering\arraybackslash}X
  | >{\centering\arraybackslash}X
  | >{\centering\arraybackslash}X
  | >{\centering\arraybackslash}m{0.1mm}
  | >{\centering\arraybackslash}m{19.85mm} 
  || >{\centering\arraybackslash}X
  | >{\centering\arraybackslash}X
  | >{\centering\arraybackslash}X|} 
    \hline
    \rowcolor{DarkBlue}\multicolumn{4}{|c|}{Binary Classification} &  {\cellcolor{white}}   & \multicolumn{4}{c|}{Multi-Class Classification}\\
    \hhline{|*4{-}|*1{~}|*4{-}|}
    \rowcolor{LightBlue} \diagbox[height=5mm, width=24mm]{\tiny Method}{\vspace{3mm}{\tiny \vspace{-1.0mm}Dominant Modality \hspace{-2.8mm}}} & Uni-Modal & Bi-Modal & Tri-Modal & {\cellcolor{white}}\multirow{-2}{*}{} & \diagbox[height=5mm, width=24mm]{\tiny Method}{\vspace{3mm}{\tiny \vspace{-1.0mm}Dominant Modality \hspace{-2.8mm}}} & Uni-Modal & Bi-Modal & Tri-Modal \\
    % \hhline{|>{\colorlb}->{\arb} ||*6{>{\arb}->{\arb}}|}
    \hline\hline 
    
    CM-FL & $\simeq$~60.12 & $\simeq$~60.56 & $\simeq$~62.25 &
    {\cellcolor{white}} &
    CM-FL & $\simeq$~46.59 & $\simeq$~47.62 & $\simeq$~48.85 \\
    \hhline{|*4{-}|*1{~}|*4{-}|}
    
    \rowcolor{LightBlue}
    CM-FL~+~DGB & $\simeq$~67.14 & $\simeq$~68.23 & $\simeq$~70.83 &
    {\cellcolor{white}} &
    CM-FL~+~DGB & $\simeq$~54.27 & $\simeq$~55.61 & $\simeq$~58.25 \\
    
    \hhline{|*4{-}|*1{~}|*4{-}|}
    
    Our Method & $\simeq$~\textbf{72.75} & $\simeq$~\textbf{73.21} & $\simeq$~\textbf{74.82} &  {\cellcolor{white}} &
    Our Method & $\simeq$~\textbf{66.43} & $\simeq$~\textbf{67.82} & $\simeq$~\textbf{69.17} \\
    
    \hline
\end{tabularx}
}
\end{table*}

\begin{table*}[ht]
\scriptsize
\caption{Global model accuracy (\%) for each dominant modality scenario for both binary classification and multi-class classification cases in the presence of class-based heterogeneity in local institution datasets.}
\vspace{-1.5mm}
\centering
\label{tab:non-uniform-level-class-based}
{
\begin{tabularx}{\textwidth}{ 
  | >{\centering\arraybackslash}m{19.85mm} 
  || >{\centering\arraybackslash}X
  | >{\centering\arraybackslash}X
  | >{\centering\arraybackslash}X
  | >{\centering\arraybackslash}m{0.1mm}
  | >{\centering\arraybackslash}m{19.85mm} 
  || >{\centering\arraybackslash}X
  | >{\centering\arraybackslash}X
  | >{\centering\arraybackslash}X|} 
    \hline
    \rowcolor{DarkBlue}\multicolumn{4}{|c|}{Binary Classification} &  {\cellcolor{white}}   & \multicolumn{4}{c|}{Multi-Class Classification}\\
    \hhline{|*4{-}|*1{~}|*4{-}|}
    \rowcolor{LightBlue} \diagbox[height=5mm, width=24mm]{\tiny Method}{\vspace{3mm}{\tiny \vspace{-1.0mm}Dominant Modality \hspace{-2.8mm}}} & Uni-Modal & Bi-Modal & Tri-Modal & {\cellcolor{white}}\multirow{-2}{*}{} & \diagbox[height=5mm, width=24mm]{\tiny Method}{\vspace{3mm}{\tiny \vspace{-1.0mm}Dominant Modality \hspace{-2.8mm}}} & Uni-Modal & Bi-Modal & Tri-Modal \\
    % \hhline{|>{\colorlb}->{\arb} ||*6{>{\arb}->{\arb}}|}
    \hline\hline 
    
    CM-FL & $\simeq$~62.34 & $\simeq$~63.02 & $\simeq$~65.42 & {\cellcolor{white}} & CM-FL & $\simeq$~49.75 & $\simeq$~51.09 & $\simeq$~53.29 \\
    
    \hhline{|*4{-}|*1{~}|*4{-}|}
    
    \rowcolor{LightBlue}  CM-FL~+~DGB & $\simeq$~67.43 & $\simeq$~68.65 & $\simeq$~71.20 &  {\cellcolor{white}} & CM-FL~+~DGB & $\simeq$~60.13 & $\simeq$~64.14 & $\simeq$~67.84 \\
    
    \hhline{|*4{-}|*1{~}|*4{-}|}
    
    Our Method & $\simeq$~\textbf{75.30} & $\simeq$~\textbf{75.83} & $\simeq$~\textbf{76.63} &  {\cellcolor{white}} & Our Method & $\simeq$~\textbf{67.17} & $\simeq$~\textbf{68.18} & $\simeq$~\textbf{70.32} \\
    
    \hline
\end{tabularx}
}
\end{table*}

We consider the first baseline to measure the system's performance under the conventional methods (i.e., FedAvg aggregation and local training using SGD), which is the naïve extension of FL to multi-modal setting as described in~\cite{XIONG2022110}. We will demonstrate that this approach will perform poorly over unbalanced modalities across the institutions. 
The second baseline~\cite{mm-fl-gradblend} implements a federated version of gradient blending introduced in~\cite{gradient-blending}, which is not tailored to address the problem of uneven number of modalities across institutions. This baseline also relies on tuning the global model aggregations weights, which is different from our method.
Further, our method removes the necessity to split the training data at each institution, as done in~\cite{gradient-blending}, to calculate the gradient blending weights. This is reflected in our formulation \eqref{equ:new-ogr-bars} where we aggregate all of the loss values needed to calculate the DGB parameters based on the modality combination of the institutions to get a distributed calculation of these parameters. Avoiding splitting the dataset in data-scarce scenarios such as ours, and generally, most medical applications, can (later shown in Sec. \ref{subsubseq:methods-compared}) improve the outcome of the federated learning. We further complement our method with PCW, which as will be shown later will give us performance edge.
In the third baseline, we aim to measure the base performance of our proposed DGB and demonstrate the performance gain we get by employing the method. Although we use this method as a baseline, we will later show that this method outperforms the state-of-the-art, e.g.,~\cite{XIONG2022110,mm-fl-gradblend}, due to consideration of different convergence speeds across various modalities. 

\vspace{-4mm}
\subsection{Hyperparamters Tuning}\label{subsec:hyperparameters}

The major hyperparameters tuned in our simulations are the number of layers for the mRNA modality's encoder depth, the initial learning-rate and its decay-rate. For the mRNA modality's encoder's depth, we conducted three experiments with three neural network architectures with multiple layer depths which are listed in Table \ref{tab:mrna-model-depth}. For the initial learning-rate, we chose five values to start the ML training with $\{10^{-2}, 10^{-3}, 10^{-4}, 10^{-5}, 10^{-6}\}$. The decay-rates for the learning-rate were searched between the values $\{0.99, 0.95, 0.90, 0.85\}$. The best performance for different mixtures of these parameters was observed for the first $50$ global FL steps and the \textit{medium} depth, $10^{-4}$ initial learning rate, and $0.99$ decay rate over each global round were chosen. We used $K=20$ local SGD iterations on each institution's ML model. Also, dividing the datasets mentioned in Sec. \ref{subsec:data}, we simulated $9$ institutions with different subsets of modalities.

\begin{figure*}[t]
\vspace{-3mm}
    % \centering
    \hspace{-0.1mm}
    \includegraphics[width=\textwidth]{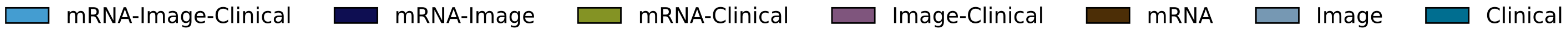}\\
    \vspace{2mm}
    \subcaptionbox{\label{subfig:rho-plot-binary-type} Binary Classification, Type-Based Heterogeneity}[0.49\textwidth]{%
        \includegraphics[width=0.49\textwidth]{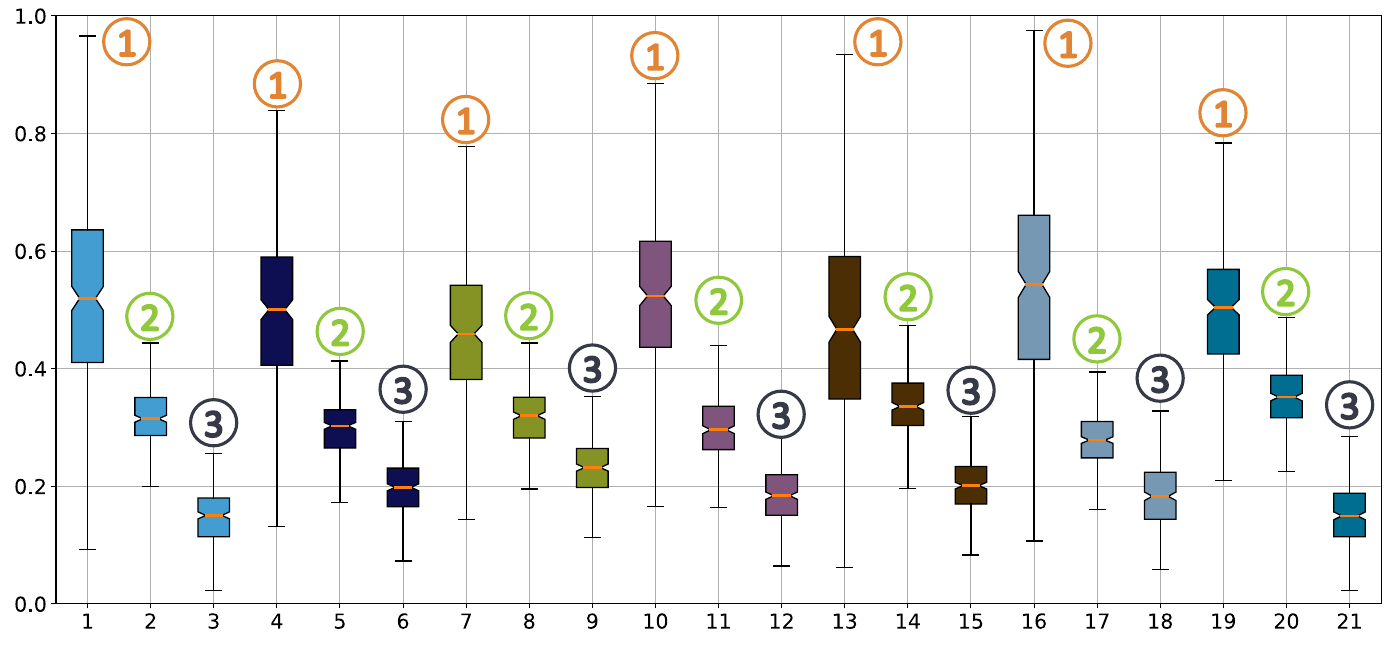}
      }
      \hfill
      \subcaptionbox{\label{subfig:rho-plot-mc-type} Multi-Class Classification, Type-Based Heterogeneity}[0.49\textwidth]{%
        \includegraphics[width=0.49\textwidth]{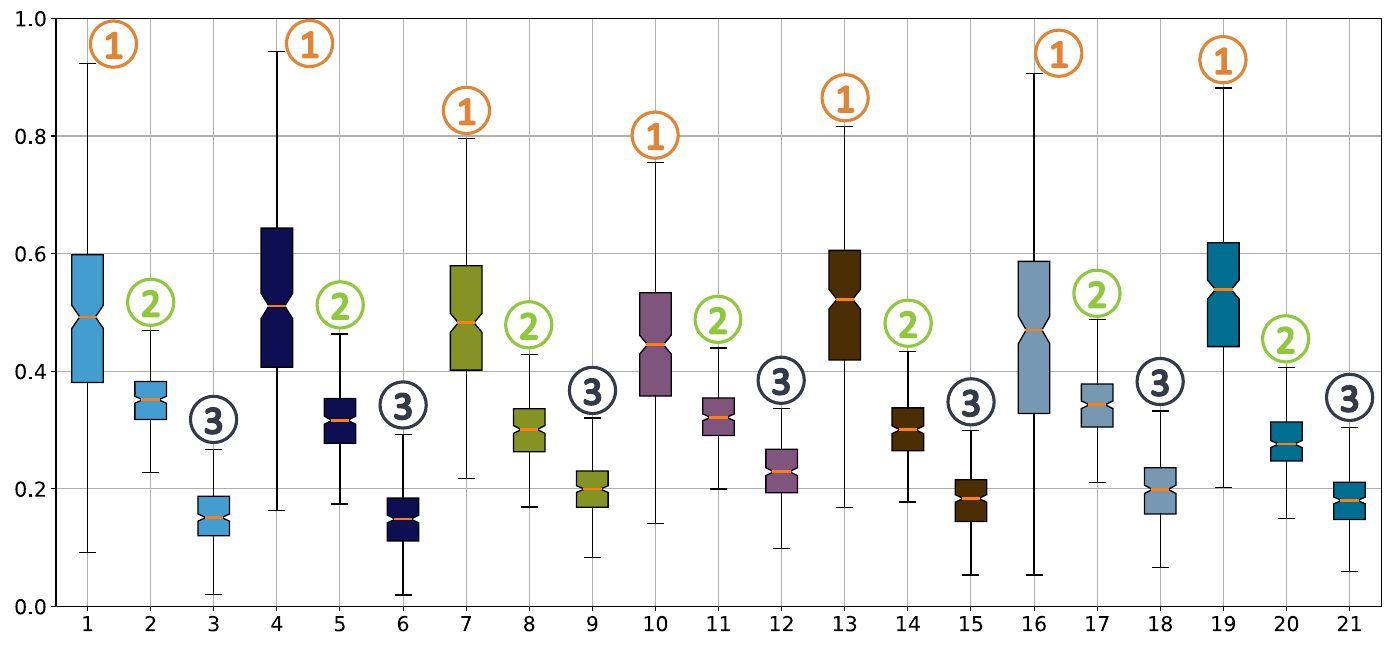}
      }\\
      
      % \vspace{2mm} % Adjust vertical spacing between subfigures as needed
      
      \subcaptionbox{\label{subfig:rho-plot-binary-class} Binary Classification, Class-Based Heterogeneity}[0.49\textwidth]{%
        \includegraphics[width=0.49\textwidth]{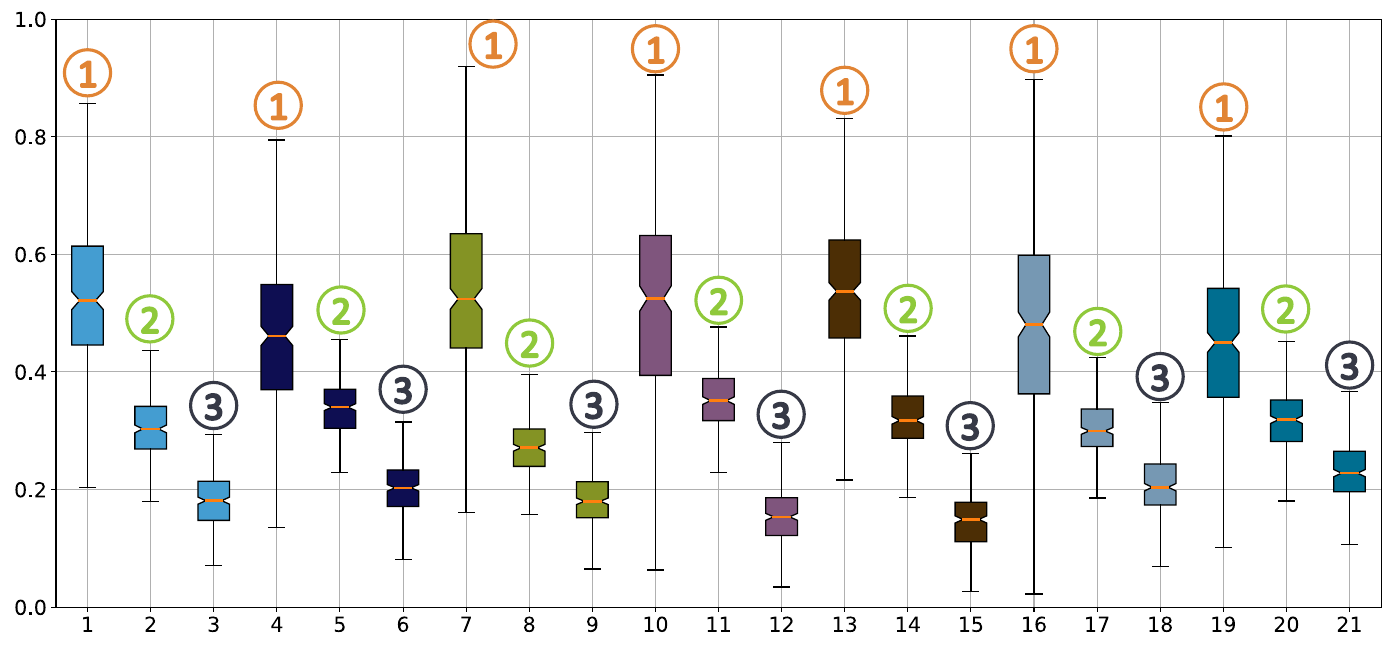}
      }
      \hfill
      \subcaptionbox{\label{subfig:rho-plot-mc-class} Multi-Class Classification, Type-Based Heterogeneity}[0.49\textwidth]{%
        \includegraphics[width=0.49\textwidth]{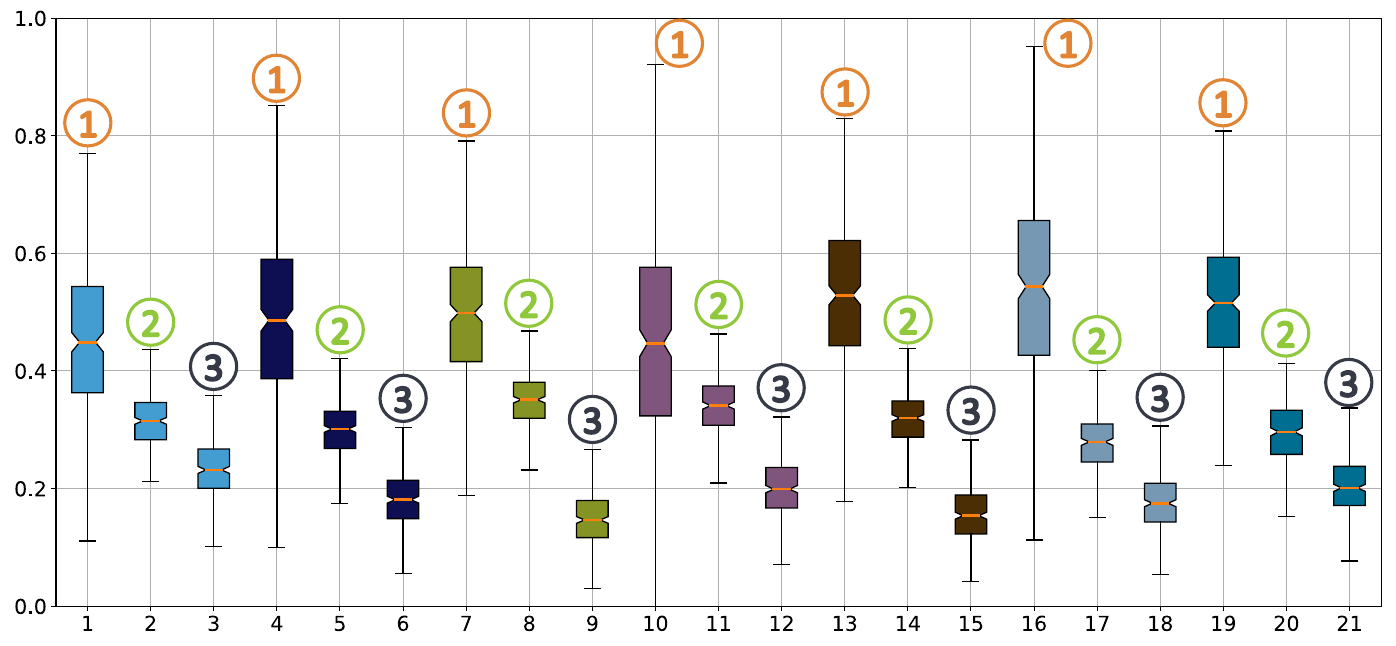}
      }
    \vspace{-2mm}
    \caption{Bar plot of the values of $\overline{\rho}_{C,n}^{(t)}$ (y-axis) for all institutions $n \in \mathcal{N}$ (x-axis) for each scenario: (a) binary classification, type-based, (b) binary classification, class-based, (c) multi-class classification, type-based, and (d) multi-class classification, class-based. The numbers beside each bar plot indicate the category of the institution's data based on Table \ref{tab:type-heterogeneity-distr} for (a) and (c) and Table \ref{tab:class-heteogeneity-distr} for (b) and (d).}
    \label{fig:rho-plots}
    % \vspace{-3.5mm}
\end{figure*}

\vspace{-2mm}
\subsection{Results and Discussions} \label{subsec:results}

We next present our simulation results and their implications.

\subsubsection{ML training under different learning methods}\label{subsubseq:methods-compared}

To test the performance of our method against the baselines, we devised an experiment trying to simulate the methods under the same circumstances. In particular,  all of the ML models start from the same initial point/model in order to avoid having random initial weights affecting the performance of the models.
For each heterogeneity case (i.e., type-based and class-based), we used the simulated institutions with the distributions in Table \ref{tab:type-heterogeneity-distr} and \ref{tab:class-heteogeneity-distr}. 
The results in Tables \ref{tab:accuracy-table-type-based} and \ref{tab:accuracy-table-class-based} show the evolution of performance of each method over time in both binary and multi-class cases for type-based and class-based heterogeneity scenarios, respectively. 
Focusing on Table \ref{tab:accuracy-table-type-based}, it is shown 
% that our method even without the addition of PCW (i.e., \textit{CM-FL + DGB}) already has a better performance than \textit{CM-FL}. The results also show 
that the  addition of \textit{DGB} initially results in a better performance, from the other methods. However, since it does not explicitly take into account the heterogeneity of data across modality combinations, it fails to retain improvement in global model performance in the later stages of training. This reveals that the client weighting method used in PCW can effectively circumvent the local ML model bias caused by non-iid data across the institutions, especially as the ML training proceeds. This observation is consistent for both binary and multi-class classification scenarios presented in the table, where our method reaches the closest performance to that of the upper-bound (i.e., UM-FL) in the later stages of training.
The table further shows that, due to the unique challenges introduced by the unbalanced distribution of data modalities and labels, baselines that are not designed to handle such non-uniformities (CM-FL, HGB-FL) fail to train the model effectively: CM-FL saturates early and starts showing overfitting behavior, while HGB-FL exhibits the same behavior later in the training.
The same behavior is observed in Table \ref{tab:accuracy-table-class-based}, where class-based heterogeneity is present across institutions' local datasets. Comparing the results between Table \ref{tab:accuracy-table-type-based} and \ref{tab:accuracy-table-class-based}, it can be seen that the methods show a slightly lower performance under type-based heterogeneity (i.e., Table \ref{tab:accuracy-table-type-based}) compared to class-based heterogeneity (i.e., Table \ref{tab:accuracy-table-class-based}).
This is due to the fact that, as discussed in Sec.~\ref{subsec:data}, type-based heterogeneity is a more fundamental level of data heterogeneity, imposing which will lead to more pronounced impact/drop on the performance.

\subsubsection{ML training under different levels of non-uniformity across the modalities}\label{subsubseq:non-uniformities-compared}
Since we are among the first to study an FL system with an unbalanced number of modalities across clients/institutions, we study the impact of varying the  non-uniformity of the number of modalities across institutions.
We compare the performance of our method against the baselines under different levels of non-uniformity to show the robustness of our method against the distribution of  modalities across institutions. 
 Since we have three modalities, we can have either uni-modal institutions, bi-modal institutions, or tri-modal institutions. We conduct our comparisons in  three settings where most institutions possess the \textit{dominant} number of modalities:
\begin{enumerate}[leftmargin=4.5mm]
    \item \textit{Uni-modal dominant:} $2/3$ of institutions have uni-modal data, $1/6$ of institutions have bi-modal data, and  $1/6$ of institutions have tri-modal data.
    \item \textit{Bi-modal dominant:} $1/6$ of institutions have uni-modal data, $2/3$ of institutions have bi-modal data, and $1/6$ of institutions have tri-modal data.
    \item \textit{Tri-modal dominant:} $1/6$ of institutions have uni-modal data, $1/6$ of institutions have bi-modal data, and $2/3$ of institutions have tri-modal data.
\end{enumerate}

\begin{table*}[ht]
\notsotiny
\caption{Accuracy (\%) of different methods under different subsets of modalities for both binary classification and multi-class classification cases in the presence of type-based heterogeneity.}
\vspace{-1.5mm}
\centering
\label{tab:accuracy-modality-type}
{
\begin{tabularx}{\textwidth}{ 
  | >{\centering\arraybackslash}X
  || >{\centering\arraybackslash}X
  | >{\centering\arraybackslash}X
  | >{\centering\arraybackslash}X
  | >{\centering\arraybackslash}X
  | >{\centering\arraybackslash}m{0.1mm}
  | >{\centering\arraybackslash}X
  || >{\centering\arraybackslash}X
  | >{\centering\arraybackslash}X
  | >{\centering\arraybackslash}X
  | >{\centering\arraybackslash}X|} 
  
    \hline
    \rowcolor{DarkBlue}\multicolumn{5}{|c|}{Binary Classification} &  {\cellcolor{white}}   & \multicolumn{5}{c|}{Multi-Class Classification}\\
    \hhline{|*5{-}|*1{~}|*5{-}|}
   \rowcolor{LightBlue} {\vspace{-10pt}}\diagbox[height=6.5mm, width=17mm]{Method}{{ Subset \hspace{-2mm}}} & \{mRNA, Image, Clinical\} & \{mRNA, Image\} & \{mRNA, Clinical\} & \{Image, Clinical\} 
    & {\cellcolor{white}}\multirow{-2}{*}{} &
    {\vspace{-10pt}}\diagbox[height=6.5mm, width=17mm]{Method}{{ Subset \hspace{-2mm}}} & \{mRNA, Image, Clinical\} & \{mRNA, Image\} & \{mRNA, Clinical\} & \{Image, Clinical\} \\ 
    \hline   \hline 
    
   CM-FL & $\simeq$~61.80 & $\simeq$~61.25 & $\simeq$~55.32 & $\simeq$~50.43 &
   {\cellcolor{white}} &
   CM-FL & $\simeq$~48.12 & $\simeq$~47.72 & $\simeq$~46.48 & $\simeq$~46.25 \\
    \hhline{|*5{-}|*1{~}|*5{-}|}
    
    \rowcolor{LightBlue}
    CM-FL + DGB & $\simeq$~69.21 & $\simeq$~65.23 & $\simeq$~55.81 & $\simeq$~54.43 &
    {\cellcolor{white}} &
    CM-FL + DGB & $\simeq$~53.83 & $\simeq$~50.18 & $\simeq$~48.43 & $\simeq$~47.69\\
    \hhline{|*5{-}|*1{~}|*5{-}|}
    
    Our Method & $\simeq$~\textbf{74.16} & $\simeq$~\textbf{68.42} & $\simeq$~\textbf{58.12} & $\simeq$~\textbf{55.84} &
    {\cellcolor{white}} &
    Our Method & $\simeq$~\textbf{67.84} & $\simeq$~\textbf{66.65} & $\simeq$~\textbf{60.72} & $\simeq$~\textbf{60.44} \\
    \hline\hline
    
    \rowcolor{LightBlue}
    Centralized (Upperbound) & $\simeq$ 87.64 & $\simeq$ 80.43 & $\simeq$ 74.13 & $\simeq$ 72.81 &
    \cellcolor{white} & 
    Centralized (Upperbound) & $\simeq$ 80.39 & $\simeq$ 76.84 & $\simeq$ 75.26 & $\simeq$ 74.32 \\
    \hline
\end{tabularx}
}
\end{table*}

\begin{table*}[ht]
\notsotiny
\caption{Accuracy (\%) of different methods under different subsets of modalities for both binary classification and multi-class classification cases in the presence of class-based heterogeneity.}
\vspace{-1.5mm}
\centering
\label{tab:accuracy-modality-class}
{
\begin{tabularx}{\textwidth}{ 
  | >{\centering\arraybackslash}X
  || >{\centering\arraybackslash}X
  | >{\centering\arraybackslash}X
  | >{\centering\arraybackslash}X
  | >{\centering\arraybackslash}X
  | >{\centering\arraybackslash}m{0.1mm}
  | >{\centering\arraybackslash}X
  || >{\centering\arraybackslash}X
  | >{\centering\arraybackslash}X
  | >{\centering\arraybackslash}X
  | >{\centering\arraybackslash}X|}
  
    \hline
    \rowcolor{DarkBlue}\multicolumn{5}{|c|}{Binary Classification} &  {\cellcolor{white}}   & \multicolumn{5}{c|}{Multi-Class Classification}\\
    
    \hhline{|*5{-}|*1{~}|*5{-}|}
    
    \rowcolor{LightBlue} {\vspace{-10pt}}\diagbox[height=6.5mm, width=17mm]{Method}{{ Subset \hspace{-2mm}}} & \{mRNA, Image, Clinical\} & \{mRNA, Image\} & \{mRNA, Clinical\} & \{Image, Clinical\} 
    & {\cellcolor{white}}\multirow{-2}{*}{} &
    {\vspace{-10pt}}\diagbox[height=6.5mm, width=17mm]{Method}{{ Subset \hspace{-2mm}}} & \{mRNA, Image, Clinical\} & \{mRNA, Image\} & \{mRNA, Clinical\} & \{Image, Clinical\} \\
    
    \hline   \hline 
    
    CM-FL & $\simeq$~65.24 & $\simeq$~63.77 & $\simeq$~53.22 & $\simeq$~52.14 &
    {\cellcolor{white}} &
    CM-FL & $\simeq$~49.49 & $\simeq$~47.98 & $\simeq$~46.97 & $\simeq$~46.46\\
    
    \hhline{|*5{-}|*1{~}|*5{-}|}
    
    \rowcolor{LightBlue}
    CM-FL + DGB & $\simeq$~71.13 & $\simeq$~69.82 & $\simeq$~56.12 & $\simeq$~54.82 &
    {\cellcolor{white}} &
    CM-FL + DGB & $\simeq$~57.07 & $\simeq$~54.04 & $\simeq$~49.49 & $\simeq$~46.97\\
    
    \hhline{|*5{-}|*1{~}|*5{-}|}
    
    Our Method & $\simeq$~\textbf{76.31} & $\simeq$~\textbf{70.21} & $\simeq$~\textbf{60.14} & $\simeq$~\textbf{57.17} &
    {\cellcolor{white}} &
    Our Method & $\simeq$~\textbf{69.92} & $\simeq$~\textbf{66.75} & $\simeq$~\textbf{61.42} & $\simeq$~\textbf{60.74} \\
    
    \hline\hline
    \rowcolor{LightBlue}
    Centralized (Upperbound) & $\simeq$ 87.64 & $\simeq$ 80.43 & $\simeq$ 74.13 & $\simeq$ 72.81 &
    \cellcolor{white} & 
    Centralized (Upperbound) & $\simeq$ 80.39 & $\simeq$ 76.84 & $\simeq$ 75.26 & $\simeq$ 74.32 \\
    \hline
\end{tabularx}
}
\end{table*}

We conduct experiments for both binary and multi-class classification cases, considering both type-based and class-based heterogeneities. The results are presented in Tables \ref{tab:non-uniform-level-type-based}, and \ref{tab:non-uniform-level-class-based}. Focusing on type-based heterogeneity (i.e., Table \ref{tab:non-uniform-level-type-based}), it can be seen that our method outperforms the other baselines by a notable margin. Also, reducing the number of modalities per institution (i.e., majority of tri-modal institutions vs. majority of uni-modal institutions) leaves a marginal impact on the performance of our method (i.e., $2.07\%$ in binary case and $2.74\%$ in multi-class case) as compared to baselines (being $2.13\%$ and $3.69\%$ in binary case and $2.26\%$ and $3.98\%$ in multi-class case). This implies the robustness of our method against the level of heterogeneity of the modalities across institutions.
The same deduction can be made from inspecting the results in Table \ref{tab:non-uniform-level-class-based}, focusing on the class-based heterogeneity. Similar to the observation made in Sec.~\ref{subsubseq:methods-compared}, comparing the results in Table \ref{tab:non-uniform-level-type-based} and \ref{tab:non-uniform-level-class-based} reveals a slight decrease in the performance under type-based heterogeneity which can be explained by the more pronounced impact of the type-based heterogeneity on ML model training as discussed in Sec. \ref{subsec:data}.

% We compare the performance (in terms of prediction accuracy) of our method against the baselines in Table \ref{tab:non-uniform-type-based}. It can be seen that our method outperforms the other baselines by a notable margin. Also, reducing the number of modalities per client (i.e., having all tri-modal clients vs. having all uni-modal clients) leaves a marginal impact on the performance of our method (i.e., $1.33\%$) as compared to baselines, being  $1.76\%$ and $2.13\%$. This implies the robustness of our method against the level of heterogeneity of the modalities across institutions.

\subsubsection{ML training under different contributing of modalities involved}\label{subsubseq:modalities-involved-compared}
In addition to heterogeneity of number of modalities across the system, the richness of data of each modality is different across institutions (i.e., some modalities may contribute more to the overall performance of the ML model as compared to the rest).
To investigate this, 
we train the ML models and gather the accuracy of the global model under availability of different subsets of modalities across institutions. We show the results of these experiments conducted under type-based heterogeneity in Table \ref{tab:accuracy-modality-type} and under class-based heterogeneity in Table \ref{tab:accuracy-modality-class}. The results in both tables under both binary and multi-class classification cases show that our method outperforms the baselines under various data modality richness levels. Also, the clinical data shows the lowest quality data as it leads to the lowest performance when combined with other modalities, while the mRNA modality has the highest quality data. This observation can further open the door to research on multi-modal FL in the presence of different quality data and attention mechanisms, which can put non-uniform emphasis on different types of data.
\begin{table*}[!h]
\scriptsize
\caption{Final global model accuracy (\%) across various cohorts' data for models trained under type-based heterogeneity.}
\vspace{-1.5mm}
\centering
\label{tab:final-acc-type-based}
{
\begin{tabularx}{\textwidth}{ 
  | >{\centering\arraybackslash}m{19.85mm} 
  || >{\centering\arraybackslash}X
  | >{\centering\arraybackslash}X
  | >{\centering\arraybackslash}X
  | >{\centering\arraybackslash}m{0.1mm}
  | >{\centering\arraybackslash}m{19.85mm} 
  || >{\centering\arraybackslash}X
  | >{\centering\arraybackslash}X
  | >{\centering\arraybackslash}X|} 
    \hline
    \rowcolor{DarkBlue}\multicolumn{4}{|c|}{Binary Classification} &  {\cellcolor{white}}   & \multicolumn{4}{c|}{Multi-Class Classification}\\
    \hhline{|*4{-}|*1{~}|*4{-}|}
    \rowcolor{LightBlue} \diagbox[height=5mm, width=24mm]{\tiny Method}{\vspace{3mm}{\tiny \vspace{-1.0mm}Cohort \hspace{-1mm}}} & BRCA & LUSC & LIHC & {\cellcolor{white}}\multirow{-2}{*}{} & \diagbox[height=5mm, width=24mm]{\tiny Method}{\vspace{3mm}{\tiny \vspace{-1.0mm}Cohort\hspace{-1mm}}} & BRCA & LUSC & LIHC \\
    % \hhline{|>{\colorlb}->{\arb} ||*6{>{\arb}->{\arb}}|}
    \hline\hline 
    
    CM-FL & $\simeq$~75.71 & $\simeq$~60.71 & $\simeq$~56.00 &
    {\cellcolor{white}} &
    CM-FL & $\simeq$~54.78 & $\simeq$~46.34 & $\simeq$~43.90 \\
    
    \hhline{|*4{-}|*1{~}|*4{-}|}
    
    \rowcolor{LightBlue}
    HGB-FL & $\simeq$~71.42 & $\simeq$~67.86 & $\simeq$~68.00 &
    {\cellcolor{white}} &
    HGB-FL & $\simeq$~72.85 & $\simeq$~53.66 & $\simeq$~56.10 \\
    
    \hhline{|*4{-}|*1{~}|*4{-}|}
    
    CM-FL~+~DGB & $\simeq$~71.43 & $\simeq$~68.57 & $\simeq$~68.00 &
    {\cellcolor{white}} &
    CM-FL~+~DGB & $\simeq$~61.21 & $\simeq$~51.22 & $\simeq$~48.78 \\

    \hhline{|*4{-}|*1{~}|*4{-}|}
    
    \rowcolor{LightBlue}
    Our Method & $\simeq$~\textbf{75.71} & $\simeq$~\textbf{75.00} & $\simeq$~\textbf{72.00} &
    {\cellcolor{white}} &
    Our Method & $\simeq$~\textbf{68.96} & $\simeq$~\textbf{68.29} & $\simeq$~\textbf{65.85} \\
    
    \hline
\end{tabularx}
}
\end{table*}

\begin{table*}[!h]
\scriptsize
\caption{Final global model accuracy (\%) across various cohorts' data for models trained under class-based heterogeneity.}
\vspace{-1.5mm}
\centering
\label{tab:final-acc-class-based}
{
\begin{tabularx}{\textwidth}{ 
  | >{\centering\arraybackslash}m{19.85mm} 
  || >{\centering\arraybackslash}X
  | >{\centering\arraybackslash}X
  | >{\centering\arraybackslash}X
  | >{\centering\arraybackslash}m{0.1mm}
  | >{\centering\arraybackslash}m{19.85mm} 
  || >{\centering\arraybackslash}X
  | >{\centering\arraybackslash}X
  | >{\centering\arraybackslash}X|} 
    \hline
    \rowcolor{DarkBlue}\multicolumn{4}{|c|}{Binary Classification} &  {\cellcolor{white}}   & \multicolumn{4}{c|}{Multi-Class Classification}\\
    \hhline{|*4{-}|*1{~}|*4{-}|}
    \rowcolor{LightBlue} \diagbox[height=5mm, width=24mm]{\tiny Method}{\vspace{3mm}{\tiny \vspace{-1.0mm}Cohort \hspace{-1mm}}} & BRCA & LUSC & LIHC & {\cellcolor{white}}\multirow{-2}{*}{} & \diagbox[height=5mm, width=24mm]{\tiny Method}{\vspace{3mm}{\tiny \vspace{-1.0mm}Cohort\hspace{-1mm}}} & BRCA & LUSC & LIHC \\
    % \hhline{|>{\colorlb}->{\arb} ||*6{>{\arb}->{\arb}}|}
    \hline\hline 
    
    CM-FL & $\simeq$~71.43 & $\simeq$~64.29 & $\simeq$~60.00 &
    {\cellcolor{white}} &
    CM-FL & $\simeq$~53.45 & $\simeq$~48.78 & $\simeq$~46.34 \\
    
    \hhline{|*4{-}|*1{~}|*4{-}|}
    
    \rowcolor{LightBlue}
    HGB-FL & $\simeq$~72.36 & $\simeq$~67.85 & $\simeq$~68.00&
    {\cellcolor{white}} &
    HGB-FL & $\simeq$~70.69 & $\simeq$~58.53 & $\simeq$~60.98 \\
    
    \hhline{|*4{-}|*1{~}|*4{-}|}
    
    CM-FL~+~DGB & $\simeq$~77.14 & $\simeq$~67.86 & $\simeq$~68.00 &
    {\cellcolor{white}} &
    CM-FL~+~DGB & $\simeq$~61.21 & $\simeq$~53.66 & $\simeq$~56.10 \\

    \hhline{|*4{-}|*1{~}|*4{-}|}
    
    \rowcolor{LightBlue} Our Method & $\simeq$~\textbf{80.00} & $\simeq$~\textbf{78.57} & $\simeq$~\textbf{72.00} &
    {\cellcolor{white}} &
    Our Method & $\simeq$~\textbf{70.69} & $\simeq$~\textbf{70.73} & $\simeq$~\textbf{68.29} \\
    
    \hline
\end{tabularx}
}
\end{table*}

\subsubsection{Comparing the values of similarity metric $\overline{\rho}_{C,n}^{(t)}$ during training for different levels of heterogeneity} \label{subsubseq:rho-plot}
In order to demonstrate how our proposed PCW method captures and reflects the heterogeneity of data across institutions during the learning process, we provide the bar plot of the values of the similarity metric $\overline{\rho}_{C,n}^{(t)}$ given by \eqref{equ:rho-bar} -- used in obtaining the adjusted average train and validation loss across the institutions in \eqref{equ:new-ogr-bars} -- during the model training period. 
\begin{table*}[ht]
\caption{Comparison of iid vs non-iid setting accuracy (\%) throughout the global aggregations under binary classification for type-based and class-based heterogeneity cases.}
\vspace{-1.5mm}
\centering
\label{tab:iid-niid-type-based}
{\scriptsize
\begin{tabularx}{\textwidth}{ 
  | >{\centering\arraybackslash}m{1.2cm}
  || >{\centering\arraybackslash}X
  | >{\centering\arraybackslash}X
  ? >{\centering\arraybackslash}X
  | >{\centering\arraybackslash}X
  ? >{\centering\arraybackslash}X
  | >{\centering\arraybackslash}X
  || >{\centering\arraybackslash}X|}

    \hline
     % \rowcolor{DarkBlue}\multicolumn{6}{|c|}{Binary Classification} &  { \cellcolor{white}}   & \multicolumn{6}{c|}{Multi-Class Classification}\\
    % \hhline{|~~~~~~|*1{-}|~~~~~~|}
    % \hhline{|*6{-}|*1{~}|*6{-}|}
    \rowcolor{DarkBlue}\cellcolor{gray} & \multicolumn{2}{c?}{iid Data} & \multicolumn{2}{c?}{non-iid Data, Type-based} & \multicolumn{2}{c||}{non-iid Data, Class-based} & \cellcolor{gray}\\
    \hhline{|*1{-}||*6{-}||*1{-}|}
    % \hhline{|*6{-}|*1{>{\arrayrulecolor[gray]{.8}}-}|*6{>{\arrayrulecolor[black]{.8}}-}|}
    % \hhline{|~~~~~~|*1{-}|~~~~~~|}
   \rowcolor{LightBlue}  \vspace{-2mm}\rowspacer{\diagbox[height=5.5mm, width=16.25mm]{\tiny \hspace{-1.25mm}Global Round}{\tiny Method}}
    & {\notsotiny CM~-~FL +~DGB} & {\notsotiny CM~-~FL +~DGB +~PCW} & {\notsotiny CM~-~FL +~DGB} & {\notsotiny CM~-~FL +~DGB +~PCW} & {\notsotiny CM~-~FL +~DGB} & {\notsotiny CM~-~FL +~DGB +~PCW} & {\notsotiny Centralized +~GB (Upperbound)}\\
    
    \hline
    \hline
    
    0 & $\simeq$ 52.34 & $\simeq$ 52.34 & $\simeq$ 52.34 & $\simeq$ 52.34 & $\simeq$ 50.16 & $\simeq$ 50.16 & $\simeq$ 55.14 \\
    
    \hline
    
    \rowcolor{LightBlue}
    100 & $\simeq$ 65.69 & $\simeq$ 65.24 & $\simeq$ 63.34 & $\simeq$ 61.80 & $\simeq$ 65.46 & $\simeq$ 63.48 & $\simeq$ 68.37 \\ 
    
    \hline
    
    200 & $\simeq$ 68.54 & $\simeq$ 67.82 & $\simeq$ 63.64 & $\simeq$ {64.04}  & $\simeq$ 67.18 & $\simeq$ 66.80 & $\simeq$ 72.49\\
    
    \hline
    
    \rowcolor{LightBlue}
    300 & $\simeq$ 71.77 & $\simeq$ 70.16 & $\simeq$ 64.24 & $\simeq$ 68.54 & $\simeq$ 68.12 & $\simeq$ 68.23 & $\simeq$ 76.30\\
    
    \hline
    
    400 & $\simeq$ 72.58 & $\simeq$~74.19 & $\simeq$ 66.43 & $\simeq$ 71.91 & $\simeq$ 69.57 & $\simeq$ 73.28 & $\simeq$ 81.45 \\
    
    \hline
    
    \rowcolor{LightBlue}
    500 & $\simeq$~75.80 & $\simeq$~76.61 & $\simeq$ 67.13 & $\simeq$ 74.16 & $\simeq$ 70.43 & $\simeq$ 76.24 & $\simeq$ 83.87 \\
    
    \hline
    
    600 & $\simeq$~78.23 & $\simeq$~79.03 & $\simeq$ 69.21 & $\simeq$ 74.16 & $\simeq$ 71.13 & $\simeq$ 76.31 & $\simeq$ 85.48 \\
    
    \hline

\end{tabularx}
}
\end{table*}

\begin{table*}[ht]
\caption{Comparison of iid vs non-iid setting accuracy (\%) throughout the global aggregations under multi-class classification for type-based and class-based heterogeneity cases.}
\vspace{-1.5mm}
\centering
\label{tab:iid-niid-class-based}
{\scriptsize
\begin{tabularx}{\textwidth}{ 
  | >{\centering\arraybackslash}m{1.2cm}
  || >{\centering\arraybackslash}X
  | >{\centering\arraybackslash}X
  ? >{\centering\arraybackslash}X
  | >{\centering\arraybackslash}X
  ? >{\centering\arraybackslash}X
  | >{\centering\arraybackslash}X
  || >{\centering\arraybackslash}X|}

    \hline
     % \rowcolor{DarkBlue}\multicolumn{6}{|c|}{Binary Classification} &  { \cellcolor{white}}   & \multicolumn{6}{c|}{Multi-Class Classification}\\
    % \hhline{|~~~~~~|*1{-}|~~~~~~|}
    % \hhline{|*6{-}|*1{~}|*6{-}|}
    \rowcolor{DarkBlue}\cellcolor{gray} & \multicolumn{2}{c?}{iid Data} & \multicolumn{2}{c?}{non-iid Data, Type-based} & \multicolumn{2}{c||}{non-iid Data, Class-based} & \cellcolor{gray}\\
    \hhline{|*1{-}||*6{-}||*1{-}|}
    % \hhline{|*6{-}|*1{>{\arrayrulecolor[gray]{.8}}-}|*6{>{\arrayrulecolor[black]{.8}}-}|}
    % \hhline{|~~~~~~|*1{-}|~~~~~~|}
   \rowcolor{LightBlue}  \vspace{-2mm}\rowspacer{\diagbox[height=5.5mm, width=16.25mm]{\tiny \hspace{-1.25mm}Global Round}{\tiny Method}}
    & {\notsotiny CM~-~FL +~DGB} & {\notsotiny CM~-~FL +~DGB +~PCW} & {\notsotiny CM~-~FL +~DGB} & {\notsotiny CM~-~FL +~DGB +~PCW} & {\notsotiny CM~-~FL +~DGB} & {\notsotiny CM~-~FL +~DGB +~PCW} & {\notsotiny Centralized +~GB (Upperbound)}\\
    
    \hline
    \hline
    
    0 & $\simeq$ 36.52 & $\simeq$ 36.52 & $\simeq$ 36.52 & $\simeq$ 36.52 & $\simeq$ 36.52 & $\simeq$ 36.52 & $\simeq$ 36.52 \\
    
    \hline
    
    \rowcolor{LightBlue}
    100 & $\simeq$ 56.56 & $\simeq$ 57.07 & $\simeq$ 46.26 & $\simeq$ 49.93 & $\simeq$ 46.83 & $\simeq$ 55.32 & $\simeq$ 58.59 \\ 
    
    \hline
    
    200 & $\simeq$ 65.65 & $\simeq$ 66.67 & $\simeq$ 48.25 & $\simeq$ 58.63 & $\simeq$ 53.82 & $\simeq$ 60.82 & $\simeq$ 69.70 \\
    
    \hline
    
    \rowcolor{LightBlue}
    300 & $\simeq$ 67.67 & $\simeq$ 68.69 & $\simeq$ 49.63 & $\simeq$ 64.32 & $\simeq$ 55.72 & $\simeq$ 64.17 & $\simeq$ 74.24\\
    
    \hline
    
    400 & $\simeq$ 69.19 & $\simeq$ 70.20 & $\simeq$ 52.36 & $\simeq$ 66.24 & $\simeq$ 51.93 & $\simeq$ 66.82 & $\simeq$ 77.27\\
    
    \hline
    
    \rowcolor{LightBlue}
    500 & $\simeq$ 70.71 & $\simeq$ 71.21 & $\simeq$ 53.64 & $\simeq$ 67.35 & $\simeq$ 56.72 & $\simeq$ 68.39 & $\simeq$ 78.28\\
    
    \hline
    
    600 & $\simeq$ 71.72 & $\simeq$ 72.22 & $\simeq$ 53.83 & $\simeq$ 67.84 & $\simeq$ 57.26 & $\simeq$ 69.92 & $\simeq$ 78.79\\
    
    \hline

\end{tabularx}
}
\end{table*}

\begin{table*}[ht]
\caption{The best accuracy (\%) acquired by all methods alongside the gap with our method across different data sizes used per institution for binary and multi-class classification under type-based heterogeneity.}
\vspace{-1.5mm}
\centering
\label{tab:size-eval-type}
{\scriptsize
\begin{tabularx}{\textwidth}{ 
  | >{\centering\arraybackslash}X
  || >{\centering\arraybackslash}X
  | >{\centering\arraybackslash}X
  | >{\centering\arraybackslash}X
  | >{\centering\arraybackslash}X
  | >{\centering\arraybackslash}m{0.1mm}
  | >{\centering\arraybackslash}X
  || >{\centering\arraybackslash}X
  | >{\centering\arraybackslash}X
  | >{\centering\arraybackslash}X
  | >{\centering\arraybackslash}X|} 
  
    \hline
     \rowcolor{DarkBlue}\multicolumn{5}{|c|}{Binary Classification} &  { \cellcolor{white}}   & \multicolumn{5}{c|}{Multi-Class Classification}\\
    % \hhline{|~~~~~~|*1{-}|~~~~~~|}
    \hhline{|*5{-}|*1{~}|*5{-}|}
    % \hhline{|*6{-}|*1{>{\arrayrulecolor[gray]{.8}}-}|*6{>{\arrayrulecolor[black]{.8}}-}|}
    % \hhline{|~~~~~~|*1{-}|~~~~~~|}
\rowcolor{LightBlue}
\vspace{-1.5mm}\diagbox[height=5mm, width=17mm]{\hspace{-1.8mm}\scalebox{0.8}{Dataset Size}\vspace{-0.3mm}}{\scalebox{0.8}{Method}\hspace{-1mm}} & CM-FL & HGB-FL & CM-FL + DGB & Our Method &
{ \cellcolor{white}}\multirow{-2}{*}{} &
\vspace{-1.5mm}\diagbox[height=5mm, width=17mm]{\hspace{-1.8mm}\scalebox{0.8}{Dataset Size}\vspace{-0.3mm}}{\scalebox{0.8}{Method}\vspace{1mm}} & {CM-FL} & HGB-FL & CM-FL + DGB & Our Method \\
\hline
\hline
30 & $\simeq$ 60.48 & $\simeq$ 64.52 & $\simeq$ 66.12 & $\simeq$ \textbf{70.96} &
{ \cellcolor{white}} &
30 & $\simeq$ 47.97 & $\simeq$ 51.51 & $\simeq$ 47.97 & $\simeq$ \textbf{63.13} \\

\hhline{|*5{-}|*1{~}|*5{-}|}

\rowcolor{LightBlue}
40 & $\simeq$ 63.71 & $\simeq$ 66.93 & $\simeq$ 68.54 & $\simeq$ \textbf{72.58} &
{ \cellcolor{white}} &
40 & $\simeq$ 48.48 & $\simeq$ 54.04 & $\simeq$ 50.51 & $\simeq$ \textbf{65.66} \\

\hhline{|*5{-}|*1{~}|*5{-}|}

50 & $\simeq$ 64.04 & $\simeq$ 68.64 & $\simeq$ 69.21 & $\simeq$ \textbf{74.16} &
{ \cellcolor{white}} &
50 & $\simeq$ 49.49 & $\simeq$ 57.52 & $\simeq$ 53.83 & $\simeq$ \textbf{67.84} \\

\hhline{|*5{-}|*1{~}|*5{-}|}

\rowcolor{LightBlue}
60 & $\simeq$ 64.52 & $\simeq$ 75.06 & $\simeq$ 76.61 & $\simeq$ \textbf{78.22} &
{ \cellcolor{white}} &
60 & $\simeq$ 50.51 & $\simeq$ 62.63 & $\simeq$ 59.60 & $\simeq$ \textbf{71.71} \\

\hline
  
\end{tabularx}
}
\end{table*}

\begin{table*}[ht]
\caption{The best accuracy (\%) acquired by all methods alongside the gap with our method across different data sizes used per institution for binary and multi-class classification under class-based heterogeneity.}
\vspace{-1.5mm}
\centering
\label{tab:size-eval-class}
{\scriptsize
\begin{tabularx}{\textwidth}{ 
  | >{\centering\arraybackslash}X
  || >{\centering\arraybackslash}X
  | >{\centering\arraybackslash}X
  | >{\centering\arraybackslash}X
  | >{\centering\arraybackslash}X
  | >{\centering\arraybackslash}m{0.1mm}
  | >{\centering\arraybackslash}X
  || >{\centering\arraybackslash}X
  | >{\centering\arraybackslash}X
  | >{\centering\arraybackslash}X
  | >{\centering\arraybackslash}X|} 
  
    \hline
     \rowcolor{DarkBlue}\multicolumn{5}{|c|}{Binary Classification} &  { \cellcolor{white}}   & \multicolumn{5}{c|}{Multi-Class Classification}\\
    % \hhline{|~~~~~~|*1{-}|~~~~~~|}
    \hhline{|*5{-}|*1{~}|*5{-}|}
    % \hhline{|*6{-}|*1{>{\arrayrulecolor[gray]{.8}}-}|*6{>{\arrayrulecolor[black]{.8}}-}|}
    % \hhline{|~~~~~~|*1{-}|~~~~~~|}
\rowcolor{LightBlue}
\vspace{-1.5mm}\diagbox[height=5mm, width=17mm]{\hspace{-1.8mm}\scalebox{0.8}{Dataset Size}\vspace{-0.3mm}}{\scalebox{0.8}{Method}\hspace{-1mm}} & CM-FL & HGB-FL & CM-FL + DGB & Our Method &
{ \cellcolor{white}}\multirow{-2}{*}{} &
\vspace{-1.5mm}\diagbox[height=5mm, width=17mm]{\hspace{-1.8mm}\scalebox{0.8}{Dataset Size}\vspace{-0.3mm}}{\scalebox{0.8}{Method}\vspace{1mm}} & {CM-FL} & HGB-FL & CM-FL + DGB & Our Method \\
\hline
\hline
30 & $\simeq$ 61.29 & $\simeq$ 66.93 & $\simeq$ 67.74 & $\simeq$ \textbf{72.58} &
{ \cellcolor{white}} &
30 & $\simeq$ 47.98 & $\simeq$ 52.02 & $\simeq$ 49.49 & $\simeq$ \textbf{64.65} \\

\hhline{|*5{-}|*1{~}|*5{-}|}

\rowcolor{LightBlue}
40 & $\simeq$ 62.03 & $\simeq$ 68.65 & $\simeq$ 70.16 & $\simeq$ \textbf{75.02} &
{ \cellcolor{white}} &
40 & $\simeq$ 48.48 & $\simeq$ 55.56 & $\simeq$ 51.52 & $\simeq$ \textbf{67.17} \\

\hhline{|*5{-}|*1{~}|*5{-}|}

50 & $\simeq$ 65.24 & $\simeq$ 69.15 & $\simeq$ 71.13 & $\simeq$ \textbf{76.31} &
{ \cellcolor{white}} &
50 & $\simeq$ 48.53 & $\simeq$ 63.56 & $\simeq$ 57.26 & $\simeq$ \textbf{69.92} \\

\hhline{|*5{-}|*1{~}|*5{-}|}

\rowcolor{LightBlue}
60 & $\simeq$ 66.12 & $\simeq$ 77.41 & $\simeq$ 75.80 & $\simeq$ \textbf{79.03} &
{ \cellcolor{white}} &
60 & $\simeq$ 52.02 & $\simeq$ 64.14 & $\simeq$ 61.62 & $\simeq$ \textbf{73.23} \\

\hline
  
\end{tabularx}
}
\end{table*}

We compare the values computed during the experiments on each of the heterogeneity types, i.e., the type-based and the class-based heterogeneities, for both binary-class and multi-class cases. In particular, the bar plots are obtained via observing the statistical distribution of the values of $\overline{\rho}_{C,n}^{(t)}$ over the training time $t$ for each institution $n$ (x-axis label of the figures) containing modality combination $C$ (represented in the legend of the figures).
Focusing on type-based heterogeneity, given the distributions of cohort data in Table \ref{tab:type-heterogeneity-distr}, the plots in Fig. \ref{fig:rho-plots}(a) and \ref{fig:rho-plots}(b) show the values of $\overline{\rho}_{C,n}^{(t)}$ for each institution $n \in \mathcal{N}$, where the numbers above each bar in the plots show the institutions corresponding Category in Table \ref{tab:type-heterogeneity-distr}. As can be deducted from Table \ref{tab:type-heterogeneity-distr}, datasets of institutions belonging to Category 1, closely resembles the global distribution of data in Table \ref{tab:data-dist}.\footnote{This is obtained by comparing the percentage of data points belong to each cohort in each institution in Table \ref{tab:type-heterogeneity-distr} to that of the global dataset in Table \ref{tab:data-dist}.} Also, it can be seen that the closeness of the local data distribution to that of the global one drops across Category 2 and 3. Thus, intuitively, we expect that, for each modality combination, institutions belonging to Category 1 contribute the most to the calculation of DOGR parameters, followed by institutions in the Categories 2 and 3. This is in fact what the plots in Fig. \ref{fig:rho-plots}(a) and \ref{fig:rho-plots}(b) show: for each modality combination, shown via a unified color, institutions from Category 1 have the highest values of $\rho$ followed by those belonging to Category 2 and 3. Further, since the order of the values of $\rho$ mimics the same order in data heterogeneity of the institutions, the results also validate that our proposed metric in PCW can differentiate between the \textit{levels of similarity} of local datasets of institutions.
The same deduction can be made focusing on class-based heterogeneity via inspecting the results depicted in Fig. \ref{fig:rho-plots}(c) and \ref{fig:rho-plots}(d), noting that the categorization of institutions in Table~\ref{tab:class-heteogeneity-distr} is similar to Table \ref{tab:type-heterogeneity-distr}: Category 1 closely resembles the global distribution of data in Table \ref{tab:data-dist} and the closeness of the local data distribution to that of the global one drops across Category 2 and 3.

\subsubsection{Comparing the performance of the trained models across various cancer types}
Given the co-existence of multiple cancer types in the dataset considered, we also study how the trained global model performs when applied on different cancer types in isolation. A higher difference in the test accuracy of the global model on the data of multiple cohorts indicates may further imply a more \textit{unbalanced} training. In order to measure the performance, we have taken the global model with the best validation accuracy during the training for each method, and tested it against the validation data of each cohort. The results for each case are included in Table \ref{tab:final-acc-type-based} and \ref{tab:final-acc-class-based}. It is shown that in
the majority of cases, the performance of baseline methods exhibit large variations across the cohorts. In particular, in most of the cases, the performance of the baseline methods are skewed towards the BRCA cohort. This can indicate the effect of data imbalance between BRCA and the other two existing cohorts (see Table \ref{tab:data-dist}) shifts the learning parameters toward the BRCA data characteristics and therefore biasing the global model and parameters to become more suitable for breast cancer classification.
Also, it can be seen that, our method exhibits the best performance across all the individual cohorts in both Table \ref{tab:final-acc-type-based} and \ref{tab:final-acc-class-based}. Further, although our method was not directly developed to improve the `fairness' across various cancer types, it still reduces the gap between the performance across various cohorts, indicating a more balanced learning. This further opens a new avenue of research on the development of novel techniques for multi-modal FL in the cancer staging context, which are focused to improve the \textit{learning fairness} across various cancer types.

\subsubsection{Comparing the performance of the trained models under iid and non-iid data distribution across the institutions}
In order to observe the effect of PCW in addressing data heterogeneity, we conduct an experiment where we compare the performance of the global model under homogeneous/iid data and heterogeneous/non-iid distributions. In the iid setting, we have sampled the data uniformly at random to create each institution's local dataset. In the non-iid case, we have implemented both type-based and class-based heterogeneities. The results of our experiments for both binary and multi-class classification cases are shown in Table \ref{tab:iid-niid-type-based} and \ref{tab:iid-niid-class-based}. As our results show, the implementation of DGB in an iid scenario (second column) reaches a higher accuracy from the implementation of DGB in a non-iid scenario (fourth and sixth columns) espeically as the training progresses; thus proving that non-iid-ness induces a notable performance drop in distributed/FL setups. The addition of PCW in a non-iid scenario (fifth and seventh columns) observably elevates the performance reached when compared with the same-type heterogeneity when PCW is not applied (fourth and sixth columns). This also proves that PCW addresses the data heterogeneity issue in the distributed/FL setups, given that the only difference between the two scenarios is the utilization of PCW.
The last column in each table also gives the result for the centralized scenario where gradient blending is applied, showing an upper bound on the performance where the heterogeneity of data is not present. Comparing the centralized case (last column) and the non-iid cases where PCW is not used (fourth and sixth columns), we can see how much the accuracy is degraded due to the heterogeneity of data (at most $16.27$\% in the case of binary classification in Table \ref{tab:iid-niid-type-based} and $24.96$\% in the case of multi-class classification in Table \ref{tab:iid-niid-class-based}). Comparing the results where PCW has been used in non-iid scenarios (fifth and seventh columns of these two tables) with the case that we have not used PCW (fourth and sixth columns), we find that PCW through addressing the presence of non-iid data has decreased the degradation in performance when compared to centralized training as follows:
\begin{itemize}[leftmargin=4.5mm]
    \item \textit{Binary Classification, Type-based Heterogeneity:} The gap is decreased by $4.95\%=74.16\%-69.21\%$. In particular, the performance gap to centralized training after applying the PCW is $11.32\%=85.48\%-74.16\%$ while this gap is $16.27\%=85.48\%-69.21\%$ when PCW is not applied.
    \item \textit{Binary Classification, Class-based Heterogeneity:} The gap is decreased by $5.18\%=76.31\%-71.13\%$. In particular, the performance gap to centralized training after applying the PCW is $9.17\%=85.48\%-76.31\%$ while this gap is $14.35\%=85.48\%-71.13\%$ when PCW is not applied.
    \item \textit{Multi-class Classification, Type-based Heterogeneity:} The gap is decreased by $14.01\%=67.84\%-53.83\%$. In particular, the performance gap to centralized training after applying the PCW is $10.95\%=78.79\%-67.84\%$ while this gap is $24.96\%=78.79\%-53.83\%$ when PCW is not applied.
    \item \textit{Multi-class Classification, Class-based Heterogeneity:} The gap is decreased by $12.66\%=69.92\%-57.26\%$. In particular, the performance gap to centralized training after applying the PCW is $8.87\%=78.79\%-69.92\%$ while this gap is $21.53\%=78.79\%-57.26\%$ when PCW is not applied.
\end{itemize}

\subsubsection{Comparing the performance of different methods across various data availability scenarios}

In order to gain a better understanding of the level of performance degradation that the scarcity of data imposes on each method, we conduct experiments under different numbers of datapoints available per institution.\footnote{Note that due to the size of the overall dataset, the maximum number of available datapoints per institution is $60$, which is reflected in Tables \ref{tab:size-eval-type} and \ref{tab:size-eval-class}.}  The results are presented in Tables \ref{tab:size-eval-type} and \ref{tab:size-eval-class} where the experiments have been conducted for the type-based and class-based heterogeneity (the results are obtained after $600$ global aggregation steps).  It can be seen that,
our method, which explicitly aims to address the heterogeneity of data (both at the modality combinations level and data belonging to each class label/cancer type available across the institutions), has been able to result in a superior performance regardless of the number of datapoints available at the institutions. 

It is worth mentioning that this case study further opens a research direction on the study and adaptation of methods that explicitly aim to address the scarcity of data (e.g., few shot learning \cite{wang2020generalizing}) to multi-modal FL under unbalanced presence of modalities across the participating institutions, an area which is yet to be explored.

\section{Conclusion and Future Works}
\noindent In this work, we developed the system model for multi-modal FL over institutions with non-iid data and unbalanced number of modalities.  To address disparate convergence rates across the various modalities involved in the system, we proposed DGB, weighing the gradients of different modalities in a non-uniform fashion. Further, to address the impact of non-iid data and the local model bias on the performance of DGB, we proposed PCW which takes into account the quality of data at institutions for refining the system-wide loss estimates fed  to DGB. We demonstrated that our proposed methodology can outperform the state-of-the-art methods. Our proposed DGB and PCW methods are generic and can be used in other tasks such as Alzheimer's disease prediction with data modalities of CT and PET scans, the exploration of which is left as future work. Another direction of research in this area is improving the feature extraction method used for each modality of data according to its distribution across the institution in the FL setting. Other future directions include using generative adversarial networks (GANs) to replicate the missing modalities at each institution are also worth looking into. Also, introducing distributed modality-aware attention mechanisms can be a promising research direction.

% \vspace{-3mm}

\bibliography{example_paper}
\bibliographystyle{ieeetr}

\end{document}